\documentclass[%
 reprint,
unsortedaddress,
longbibliography,
nofootinbib,
 amsmath,amssymb,
 aps,
pra,
]{revtex4-1}
\usepackage{scrextend}
\usepackage{xcolor}
\usepackage{soul}
\usepackage{graphicx}
\usepackage{dcolumn}
\usepackage{bm}
\usepackage{hyperref}

\usepackage{amsfonts,amsthm,amsmath,amssymb,graphicx,color,youngtab,bookmark,subcaption}
\newcommand{\corr}[1]{\left\langle#1\right\rangle} 
\usepackage[T1]{fontenc}
\usepackage[utf8]{inputenc}
\usepackage[sort&compress]{natbib}

\def\corr#1{\left\langle #1 \right\rangle}

\newcommand{\be}{\begin{equation}}
\newcommand{\bea}{\begin{eqnarray}}
\newcommand{\ba}{\begin{equation}\begin{aligned}}

\newcommand{\ee}{\end{equation}}
\newcommand{\eea}{\end{eqnarray}}
\newcommand{\ea}{\end{aligned}\end{equation}}

\begin{document}


\title{Why Unsupervised Deep Networks Generalize}

\author{Anita~de~Mello~Koch$^2$, Ellen~de~Mello~Koch$^2$ and Robert de Mello Koch$^{1,3}$}
\affiliation{ Guangdong Provincial Key Laboratory of Nuclear Science$^1$, Institute of Quantum Matter, 
South China Normal University, Guangzhou 510006, China}
\affiliation{Guangdong-Hong Kong Joint Laboratory of Quantum Matter$^1$, Southern Nuclear Science Computing Center, 
South China Normal University, Guangzhou 510006, China}
\affiliation{School of Electrical and Information Engineering$^2$, University of the Witwatersrand, Wits, 2050, South Africa} 
\affiliation{National Institute for Theoretical Physics and Mandelstam Institute for Theoretical Physics$^3$, University of the Witwatersrand, Wits, 2050, South Africa.}%
\email{demellokochanita@gmail.com,\\
ellendemellokoch@gmail.com,\\
robert.demellokoch@gmail.com}


\date{\today}

\begin{abstract}
Promising resolutions of the generalization puzzle observe that the actual number of parameters in a deep network is 
much smaller than naive estimates suggest.
The renormalization group is a compelling example of a problem which has very few parameters, despite the fact
that naive estimates suggest otherwise.
Our central hypothesis is that the mechanisms behind the renormalization group are also at work in deep learning, 
and that this leads to a resolution of the generalization puzzle.
We show detailed quantitative evidence that proves the hypothesis for an RBM, by showing that the
trained RBM is discarding high momentum modes. 
Specializing attention mainly to autoencoders, we give an algorithm to determine the network's parameters directly from 
the learning data set.
The resulting autoencoder almost performs as well as one trained by deep learning, and it provides an excellent
initial condition for training, reducing training times by a factor between 4 and 100 for the experiments we considered.  
Further, we are able to suggest a simple criterion to decide if a given problem can or can not be solved using a deep network. 
\end{abstract}

\maketitle

\section{Introduction}\label{Intro}

Training a deep network involves applying an algorithm, which fixes the parameters of the network, using a training 
data set as input.
After training is complete, the deep network is evaluated, by studying the trained network's performance 
on unseen test data.
The difference between how the network performs on the test data and how it performs on unseen data defines a 
generalization error.
Networks that perform as well on unseen data as they did on training data, have a small generalization error.
On the other hand, networks that do well on training data but fail on unseen data, have a large generalization error.

Even with an incomplete understanding of how deep learning works, we have definite expectations for the size 
of the generalization error, based essentially on common sense.
If the training data set is much smaller than the number of parameters in the network, training can fit the
data perfectly, so that not only trends in the data but also errors and noise are captured during training.
Intuitively we expect the generalization error is large, a penalty for fitting noise.
If the training data set is much larger than the number of parameters, training forces the network to ignore noise in the
data and only consistent and repeated trends will be learned.
In this case we expect the generalization error is small.
For the deep networks we discuss in this article, we have hundreds of millions of parameters trained using data sets
with tens of thousands of elements.
These are typical sizes of data sets and numbers of network parameters in practical deep learning applications.
Clearly then, we are squarely in the regime of large generalization errors.
Remarkably however, for typical deep learning applications, the generalization error is small.
This begs the question: why do deep nets generalize?

The fact that such a basic puzzle remains open underscores that despite the remarkable success of deep
learning applications, there is till a lot to learn about how deep learning works.
One does not need to prove generalization since it is well established in practice.
However, an understanding of why networks generalize will shed light on what properties define well trained networks
and in this way it will suggest how to find better architectures and more efficient training algorithms.
The importance of the generalization question is appreciated and it has been studied for decades.
Early attempts to explain generalization, suggested \cite{hochreiter1997flat,hinton1993keeping} 
that deep networks that have flat minima in the training loss
generalize well.
Intuitively this is rather appealing: a potential that is truly flat in a given direction does not depend on the corresponding
coordinate, and it is not a parameter of the network.
With a flat enough potential, the effective number of network parameters may be small enough that the generalization
puzzle is resolved.   
Numerical studies seem to support this idea, by demonstrating that sharp minima lead to higher generalization errors 
\cite{keskar2016large}.
The ``flatness'' of the potential can be quantified in terms of the stability of the network's output, under addition of noise
to the network's parameters \cite{langford2001not}.
By exploiting this noise stability phrasing of the problem, a number of generalization bounds were derived \cite{mcallester1999some,neyshabur2017pac,neyshabur2017exploring,bartlett2002rademacher,neyshabur2015norm,bartlett2017spectrally,neyshabur2017pac,golowich2018size},
but the bounds obtained are not yet tight enough to resolve the puzzle.
Another notion of noise stability is the networks ability to reject noise injected at its input or internal nodes of the network
\cite{morcos2018importance}.
Using this version of noise stability \cite{arora2018stronger} proved generalization bounds that are almost strong enough to
resolve the generalization puzzle.
The insights of \cite{arora2018stronger} strongly motivated the study presented in this paper, so it is worth reviewing
some of the key ideas.

The article \cite{arora2018stronger} starts off by considering a fully connected layer, meaning that each input neuron
is connected to every output neuron.
The transformation from the input to the output neurons is linear and consequently one can perform a 
singular value decomposition of the weight matrix constructed by the training algorithm.
Many of the resulting singular values are small, and can be set to zero.
Consequently these should not be counted as parameters of the deep network, so that we have a much smaller number of 
``effective parameters''.
An operational test that there are a small number of singular values is captured by studying the noise rejection properties
of the network: if noise is added to the input of a layer, it hardly has an effect on the output of that layer.
The idea behind the equivalence of noise rejection and a small number of large singular values is rather simple: noise 
projects equally onto all singular vectors, but it is only the noise that is projected onto singular vectors associated to large 
singular values that is transmitted by the network. 
If only a small fraction of the singular values are large, only a small fraction of the noise is transmitted. 
For a network with many layers, the problem is non-linear so we can not use the singular value decomposition to analyze the
network.
However, the idea of noise rejection does generalize to a multi layer network, and this is why it is useful.
For the multi layer case, we can compute a Jacobian which tells us how output changes in response to small changes in input.
Trained multi layer deep networks again exhibit noise rejection \cite{arora2018stronger}.
The observed noise rejection is evidence that the network only passes input information lying in a small subspace
and so it again indicates that we are dealing with a very special map and that only a very small number of the total
parameters that might appear in this map actually appear.
This is a powerful idea: the naively very large number of parameters translates into a much smaller number of 
effective parameters and it is only the values of these effective parameters that actually determine the output of the deep 
net\footnote{This reduction is dramatic. For the cases we study here, $\approx 0.3$\% of the parameters are effective
parameters.}.
Using this much smaller number of effective parameters we come close to understanding why deep nets generalize.

The explanation provided in \cite{arora2018stronger} is convincing, but it can only be part of the answer.
To resolve the generalization puzzle, one needs an understanding of why so many of the singular values are small.
Two natural questions to consider are: ``Is it generic that such a large set of parameters ultimately get reduced to a very much 
smaller set of effective parameters?'' and ``What is the mechanism responsible for ensuring that almost all of the parameters 
of the deep network are irrelevant, that only a handful of the parameters actually count?''
It is this class of questions that concern us in this study.
A good answer should argue that the appearance of a large number of very small singular values is generic and further 
it should provide some insight into the basic mechanism for why this happens.
Our approach to these questions shows that the structures that appear in the renormalization group descriptions 
of coarse graining in statistical physics and quantum field theory, naturally appear in deep learning.

The renormalization group is used to generate a low distance effective theory (called the infrared or IR theory) starting
from a microscopic theory (called the ultraviolet or UV theory).
The UV theory can be specified in terms of a Hamiltonian, giving the energy of the system.
Apriori, there are an enormous number of possible parameters of the theory corresponding to the possible terms that
can be added to the Hamiltonian.
Typically any term that is consistent with the symmetries of the theory can be added, so that the Hamiltonian is given
by an infinite series and each term in the series has its own coupling constant.
These coupling constants are the parameters of the UV theory.
The renormalization group classifies parameters as relevant, marginal or irrelevant, according to how they evolve under
the coarse graining. 
Relevant parameters grow as the coarse graining proceeds, irrelevant parameters decay and marginal parameters are fixed.
Further, in situations in which the coarse graining has a well defined end point, i.e. in which there is a long distance
effective description, most parameters of the problem are irrelevant.
This provides a natural setting in which a problem with a potentially enormous number of parameters turns out to have very 
few effective parameters.
Our central hypothesis in this paper is that the mechanisms behind the renormalization group are also at work in deep learning, 
and that this leads to a resolution of the generalization puzzle.

Our motivation for this study was to map the deep learning problem into the renormalization group language and thereby 
to shed light on how the generalization puzzle might be resolved.
We will see that this is a useful exercise and what is more, it suggests a new framework within which a theoretical description
of deep learning can be developed.
For the case of an RBM trained on Ising data, a study of the trained network proves that the network is coarse graining
the input data, by discarding high momentum modes.
The Ising example teaches us what to look for to recognize the renormalization group transformation, in the trained RBM.
With this experience in hand, we find that autoencoders trained on various standard data sets all perform
renormalization group transformations.
This new point of view achieves a detailed understanding of what the network is doing and precisely what it learns,
enabling us to give an algorithm to compute the network's parameters, given the learning data set.
Although the resulting network does not perform quite as well as a trained network, it is an excellent initial condition for
learning, cutting training times by a factor between 4 and 100 for the experiments we performed.
Further, we are able to suggest a simple criterion to decide if a given problem can or can not be solved 
using a deep network. 

In the next section we consider unsupervised learning using a restricted Boltzmann machine (RBM).
To engineer a deep learning setting, which is as similar to the renormalization group set up as possible, we train an
RBM using data given by states of an Ising model.
The trained weight matrix has a small number of large singular values and many small ones, which is the scenario of interest.
We give the trained weight matrix a coarse graining interpretation, by showing that it is a quantitative 
match to block spin averaging, naturally used to implement the renormalization group for spin systems.
A novel feature is that blocks overlap so that a spin in the UV lattice is averaged in more than one IR spin.
This coarse graining interpretation gives an attractive interpretation of the singular value decomposition of the weight matrix.
Singular vectors with a large singular value are relevant, and those with a small singular value are irrelevant. 
Further, the hidden singular vectors are obtained by discarding the high momentum modes of the visible singular vectors, so
that the coarse graining being performed by the RBM is an exact match for the renormalization group transformation 
close to the free field fixed point!
We go on to understand how the relevant degrees of freedom are determined, directly from the data set used for learning.
We also consider a 3-layer stacked RBM, as well as autoencoders trained on the MNIST data set and data sets
given by images of flowers and clouds.
We present our conclusions in Section \ref{Conc} and discuss the implications of thinking about deep learning in terms of a 
coarse graining perspective, instead of the more usual curve fitting paradigm.
The Appendices collect technical results and further supporting data obtained from numerical experiment.

\section{Unsupervised Learning}\label{background}

In this section we study an RBM trained on data given by the states of an Ising model, at fixed temperature on a 
rectangular lattice. 
The renormalization group interpretation for the Ising problem is well studied, so this is a good starting point.
Our primary goal is to explain why deep networks have far fewer parameters than suggested by naive estimates.
The layers that are completely connected, i.e. they have a connection between every visible and hidden neuron, 
often contain the majority of parameters of the network.  
The RBM is a completely connected layer so it is a good example of a layer with many parameters.

In subsection \ref{ising} we study a single layer RBM trained on Ising data.
This allows us to establish a link to the renormalization group.
A useful way to implement the renormalization group is through block spin transformations, which are reviewed in
Appendix \ref{sec:blockspin}.
We explore this connection both at the level of singular values and singular vectors.
Singular vectors with dimension given by the number of input neurons will be called visible singular vectors,
while singular vectors with dimension given by the number of output neurons will be called hidden singular vectors
A network with three layers is also considered.
One valid criticism of this work is that lessons learned from the Ising data might not be applicable to more
usual applications.
In subsection \ref{mnist} we train using the MNIST data set, in subsection \ref{flowers} we use a data set of images
of flowers and in \ref{clouds} one of clouds.
These examples all exhibit the general features uncovered using the Ising data set.

\subsection{Ising}\label{ising}

We trained an RBM on Ising data.
The input to the network is a set of 6400 dimensional vectors whose entries are the values ($\pm 1$) of spins located
on the sites of an 80$\times$80 rectangular lattice.
The spin states used for training were generated using a Monte Carlo simulation with the Boltzmann distribution for 
the Ising Hamiltonian at a temperature of $T=4$.
The output from the network is a set of 1600 dimensional vectors whose entries are the values ($\pm 1$) of spins located
on the sites of a 40$\times$40 rectangular lattice.
Thus, the weight matrix of the RBM has 10 240 000 entries, constituting the vast majority of the total number of
parameters of the network. 

During training we used batch sizes of 1000 samples, and training was performed with a total of 40 000 samples.
The learning rate was $1\times 10^{-3}$ and we used a $\tanh (\cdot)$ activation function. 
The total number of training samples was chosen so that the training error had converged.
See Appendix \ref{RBM} for a detailed description of the RBM including the equations we used in this study, and see 
Appendix \ref{sec:ising} for a discussion of the Ising model together with the Hamiltonian we used.

\begin{figure}[h!]
    \centering
    \includegraphics[width=0.5\textwidth]{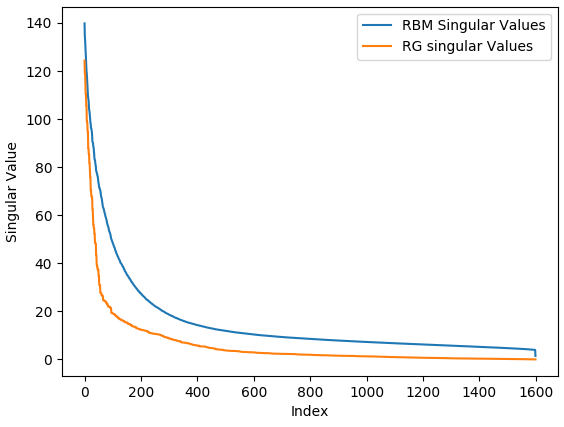}
    \caption{A comparison of the singular values of the weight matrix of the trained RBM and the singular values of the
                matrix implementing the renormalization group by block spin averaging.}
    \label{fig:isingrgcompare}
\end{figure}

The trained weight matrix defines a linear map from input neurons to output neurons.
One way to exhibit the structure of a linear map is by the singular value decomposition.
In particular, it can interrogate how many parameters the linear map actually has.
A large number of small singular values implies that there are not many parameters in the map.
The weight matrix trained using Ising data has many small singular values, and the size of the singular values has a 
distinctive fall off, described by a smooth curve.
The block spin averaging, also a linear operation, can also be studied using the singular value decomposition.
As shown in Figure \ref{fig:isingrgcompare}, we find that the plot of the singular values of the trained RBM weight matrix and
the block spin coarse graining, are almost identical.
Here we choose the size of the blocks that are averaged to produce a block spin, to maximize agreement with the 
RBM weight matrix. 
One novel feature of our results is that blocks that are averaged to produce the block spin, overlap.
In the usual block spin averaging one averages a block of spins to produce an effective spin (as we do here) and the 
averaging rule does not allow blocks to overlap (unlike what we do here), i.e. each spin that is averaged contributes to 
only one effective spin\cite{wilson1975renormalization}.
There is however precedent for overlapping blocks: correlations between blocks, which are key for successful real space
renormalization group algorithms\cite{white1998strongly}, are captured with an overlapping block 
approach \cite{degenhard2002real}.
See Appendix \ref{blockoverlap} for more details of the block spin averaging.

The real space renormalization group transformation is well understood: it is performing a coarse graining.
The singular value decomposition provides a basis on which the coarse graining has a particularly simple action:
coarse graining a visible singular vector gives the corresponding hidden singular vector multiplied by the singular value.
Very small singular values correspond to visible singular vectors that are essentially coarse grained to zero.
These are naturally associated with irrelevant degrees of freedom.
The visible singular vectors that are associated with large singular values dominate the answer of coarse graining and
hence are the relevant degrees of freedom.
Consequently, the singular value decomposition is breaking the coarse graining up according to relevant and irrelevant
degrees of freedom.

\begin{figure}[h!]
    \centering
    \includegraphics[width=0.5\textwidth]{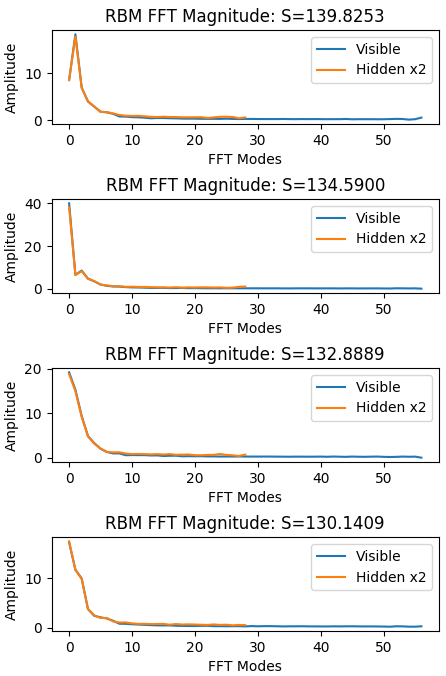}
    \caption{The Fourier transform of the visible and hidden singular vectors associated to the five largest singular vectors,
                obtained from the singular value decomposition of the trained RBM weight matrix.}
    \label{fig:singvecsisingrbm}
\end{figure}

We know that the renormalization group is constructing the low energy effective theory.
Consequently we expect the relevant degrees of freedom are associated with long distance behaviour of system.
Another way to say this, is that the Fourier transform of the relevant degrees of freedom should have its support in the
lowest Fourier modes.
We can explore the singular vectors to probe this expectation.
The singular vectors are states on the lattice.
The Fourier transform on the lattice is well defined, so we can take the Fourier transform of these states.
To produce a one dimensional plot we average the magnitude of the FFT over angles to
obtain a radial FFT. See Appendix \ref{radialFFT} for the details.
This radial average is well motivated physically since we are simply exploiting the rotational symmetry of the problem.
The rotational symmetry is broken by the lattice, so it is only an exact symmetry in the continuum limit.

\begin{figure}[h!]
    \centering
    \includegraphics[width=0.5\textwidth]{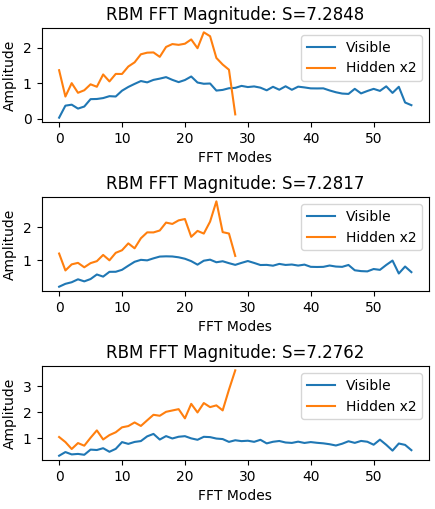}
    \caption{The Fourier transform of the visible and hidden singular vectors associated to the five smallest singular vectors,
                obtained from the singular value decomposition of the trained RBM weight matrix.}
    \label{fig:smallsingvecsisingrbm}
\end{figure}

The results for the Fourier transform of the singular vectors shown in Figure \ref{fig:singvecsisingrbm} and 
Figure \ref{fig:smallsingvecsisingrbm} are intuitively appealing.
Both the visible and hidden singular vectors associated with the large singular values have all of their support for small
Fourier modes.
Further, these hidden and visible singular vectors are identical, after a simple rescaling - multiplication by two.
The hidden singular vectors are defined on a lattice with fewer total lattice sites, so that if we change units to keep the
total length of the lattice fixed, the hidden singular vectors are obtained from the visible singular vectors by discarding
the high momentum modes. 
As Figure \ref{fig:singvecsisingrbm} shows, this is not an approximate statement - after the high momentum modes of
the visible singular vectors are discarded, there is an exact equality with the hidden vectors. 
This is remarkable because the procedure of discarding high momentum modes is an exact match to the form of the 
momentum space renormalization group transformation near the free field fixed point!
The momentum space renormalization group transformation near the free field fixed point is reviewed in 
Appendix \ref{momSpaceRG}.
Since the transformation discards modes that are zero anyway, one might mistakenly conclude that the transformation
is trivial.
This is not the case.
Indeed, the Fourier transforms associated with the hidden and visible singular vectors are different, so that in the
end the above transformation is extracting long distance features.

In contrast to this, the singular vectors associated with small singular values do not have most of their support at
low Fourier modes, but are spread across the complete Fourier space.
Further, the hidden and visible singular vectors are no longer related by the free field fixed point renormalization group.
These modes are irrelevant and should simply be dropped.
Dropping these does not affect the performance of the network.

The interpretation of the RBM in terms of coarse graining suggests that multiplying the weight matrix into a visible vector 
can be interpreted as follows:
The size of the singular value tells you how relevant the mode is.
We keep only the terms that are associated with large singular values.
Each visible singular vector associated to a large singular value defines a relevant degree of freedom.
The corresponding hidden singular vector is what the averaged relevant mode becomes after coarse graining.
The visible singular vectors that correspond to small singular values correspond to irrelevant modes. 
The weight matrix decomposes the state it acts on into relevant and irrelevant modes and the irrelevant modes are
discarded, while the relevant modes are averaged.

\begin{figure}[h!]
    \centering
    \includegraphics[width=0.5\textwidth]{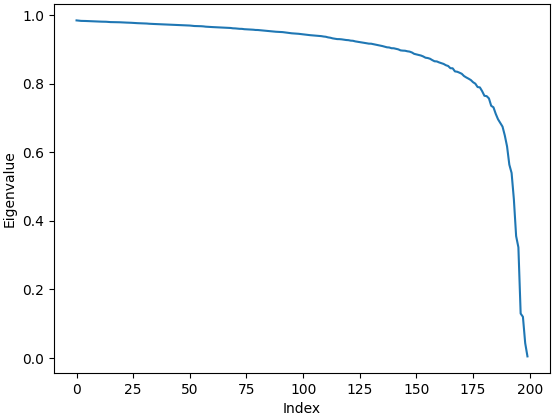}
    \caption{The eigenvalues of $O=P_{\rm data}P_{\rm trained}P_{\rm data}$  computed using the trained
                 weight matrix and the Ising training data set.}
    \label{fig:ising_check}
\end{figure}

The specific form of the relevant modes is determined by the training data set.
The fact that an effective description even exists strongly suggests that the data is concentrated on a sub manifold
of the vector space in which the data lives.
If this is the case, it is natural to expect that the visible singular vectors span this space.
This simple picture provides a definition for the relevant modes, directly in terms of the data.
A nice feature of this proposal, is that it is simple to test numerically using the training data.
Using the data, we can evaluate the data covariance matrix and study it's eigen decomposition.
If there are a few large eigenvalues and many small ones, the data is indeed concentrated on a sub manifold.
The projection operator, projecting to this sub manifold is constructed using the eigenvectors associated
to the large eigenvalues.
Call this projector $P_{\rm data}$; it's eigenvalues are, as for any projection operator, given by 0 or 1. 
We can also construct a projection operator that projects to the sub space spanned by the visible singular vectors
associated to the large singular values.
Call this projector $P_{\rm trained}$; it's eigenvalues are, again, 0 or 1. 
The operator
\bea
O=P_{\rm data}P_{\rm trained}P_{\rm data} 
\eea
will measure the alignment of these two subspaces.
If all of $O$'s eigenvalues are 0 or 1, and the number of eigenvalues equal to 1 matches the number of large singular
values, then the subspaces are perfectly aligned.
In general we expect many zero eigenvalues, plus eigenvalues that are close to, but not exactly equal to 1.
The closer these eigenvalues are to 1 the better the two subspaces are aligned. 
The numerically computed eigenvalues, shown in Figure \ref{fig:ising_check}, demonstrate that there is a convincing alignment
between the two subspaces.
They are evaluated using the singular vectors associated to the largest 200 singular values and the eigenvectors associated
to the largest 200 eigenvalues of the data covariance matrix. 
The numerical results show that of the 200 eigenvalues that might be 1, a majority of 177 of them are above 0.8 and only 
two are close to zero.
This check is non-trivial: since both subspaces are embedded in a 1600 dimensional space, there was plenty of room for
them to be completely orthogonal. 
These results are in agreement with observations reported in \cite{baldi1989neural,plaut2018principal,jing2020implicit}.

Since we have developed an understanding of the irrelevant and relevant modes of the weight matrix it makes sense to revisit 
the counting of effective parameters.
There are three important ideas 
\begin{itemize}
\item[1.] The sum over singular values can be truncated since many of the singular values are very small.
\item[2.] The visible singular vectors have support only at low Fourier modes.
High Fourier modes are not parameters as they are set to zero.
\item[3.] Since the hidden singular vector is identical to the visible singular vector, we do not
introduce new parameters for the hidden singular vector.
\end{itemize}
The reduction in the number of parameters implied by point 1 above is the reduction discussed in \cite{arora2018stronger}.
Keeping the largest 200 singular values (which amounts to dropping singular values less that 20\% of the largest) reduces 
the number of parameters to 1 280 000, which is still very much larger than the number of data samples.
For this particular example, the singular value curve is not very sharp and although this reduction in the number of
effective parameters look impressive, its a little on the low side compared to typical.
Even in more favorable cases, the reduction in the number of effective parameters achieved by the singular value 
truncation is not quite enough to resolve the generalization puzzle. 
Points 2 and 3 above are further reductions, not yet appreciated in the literature.
This new reduction only becomes apparent after taking a Fourier transform, as instructed by the renormalization group logic.
Looking at the singular vectors in Figure \ref{fig:singvecsisingrbm}, it is clear that all modes above Fourier mode 10 vanish.
The fact that we have taken a radial average implies that the number of distinct modes grows as the square of the mode
number.
Consequently, setting all modes above mode 10 to zero implies a further dramatic reduction in the number of effective 
parameters, by a factor of 36.
This leaves us with 35 555 effective parameters, which is indeed less than the number of data samples in the training
data set, explaining why we should expect the network to generalize i.e. why there is no over fitting.
Thus, truncating the smallest singular values takes us a long way, but it is a nudge from the renormalization group
that gets us over the line.
\begin{figure}[h!]
    \centering
    \includegraphics[width=0.45\textwidth]{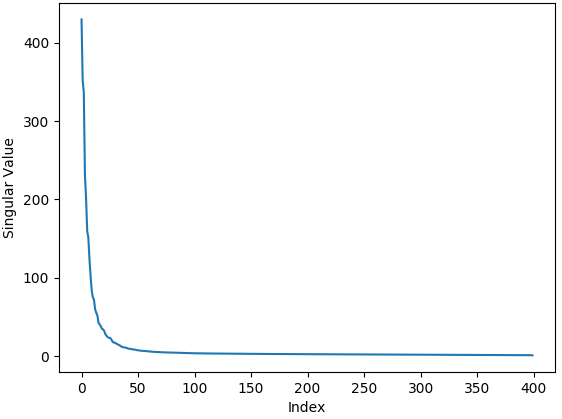}
    \caption{The singular values obtained from a singular value decomposition of the trained weight matrix of the
                  second layer of the stacked RBM.}
    \label{fig:layer2}
\end{figure}

An interesting extension of the study above is to consider stacked RBMs.
We consider a three layer network.
The first layer is as above, that is, the input is given by a set of states of an $80\times 80$ lattice and the output is
a set of states of a $40\times 40$ lattice; the second layer outputs states of a $20\times 20$ lattice and the third
states of a $10\times 10$ lattice.
Training uses the input data set described above.
Training is carried out in a greedy layer-wise fashion.

\begin{figure}[h!]
    \centering
    \includegraphics[width=0.45\textwidth]{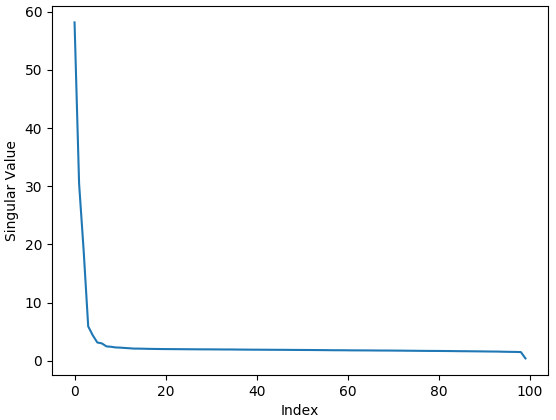}
    \caption{The singular values from a singular value decomposition of the trained weight matrix of the
                  third layer of the stacked RBM, trained using Ising data.}
    \label{fig:layer3}
\end{figure}

The singular value curves are shown in Figure \ref{fig:layer2} and in Figure \ref{fig:layer3}.
It is clear that the singular value curves go to zero ever more steeply as the depth increases.
Further, we have checked that the Fourier transform of the visible and hidden vectors are again in complete
quantitative agreement, up to the scaling by 2 we found in the first layer.
This is again a smoking gun for momentum space renormalization group coarse graining near the free
field fixed point.
The reader interested in these numerical results is referred to Appendix \ref{sec:numeric}.
Consequently, our conclusions continue to hold when we increase the number of layers, to obtain a deep network.
This is completely unsurprising, since the network is trained in a greedy layer-wise fashion.

\subsection{MNIST}\label{mnist}

A reasonable criticism of the previous section is that Ising data is rather special, and perhaps more closely related
to usual renormalization group applications than the typical deep learning data set.
For that reason, we consider the MNIST data set \cite{lecun-mnisthandwrittendigit-2010} in this section.
The input to the RBM is images containing $28\times 28$ pixels and the output from the RBM are images containing
$14\times 14$ pixels. 
\begin{figure}[h!]
    \centering
    \includegraphics[width=0.5\textwidth]{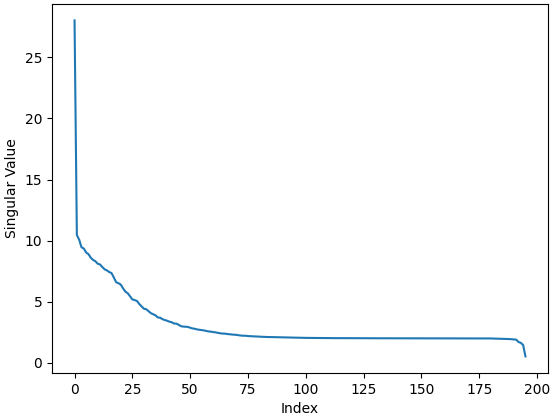}
    \caption{The singular values obtained from a singular value decomposition of the trained weight matrix of an RBM
                 trained using the MNIST data set.}
    \label{fig:mnist_sing}
\end{figure}

\begin{figure}[h!]
    \centering
    \includegraphics[width=0.5\textwidth]{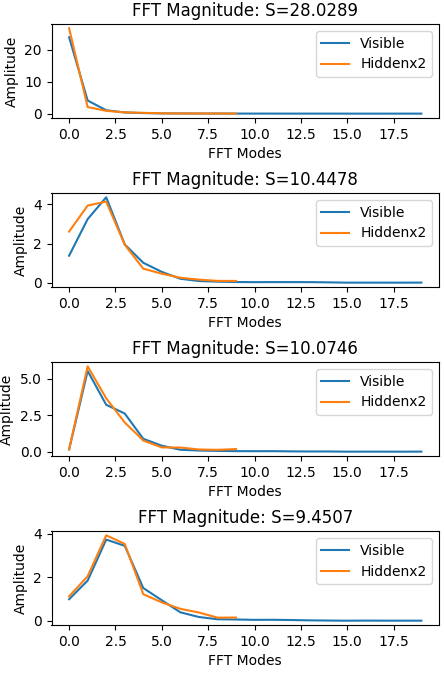}
    \caption{The radial average of the Fourier transform of the singular vectors associated to the largest singular values, of the
                 RBM weight matrix trained on the MNIST data set.}
    \label{fig:mnist_sing_vec}
\end{figure}

The singular values of the trained weight matrix are shown in Figure \ref{fig:mnist_sing}.
There are again a large number of vanishing singular values, which may be set to zero.
In the previous subsection, the strongest evidence, proving that the RBM is performing a
renormalization group coarse graining, is given by verifying that the hidden singular vectors are given by discarding
the high momentum modes of the visible singular vectors.
The radially averaged Fourier transform of the singular vectors obtained from the MNIST data set are shown in 
Figure \ref{fig:mnist_sing_vec}, proving that the RBM is again carrying out a renormalization group transformation.

\begin{figure}[h!]
    \centering
    \includegraphics[width=0.5\textwidth]{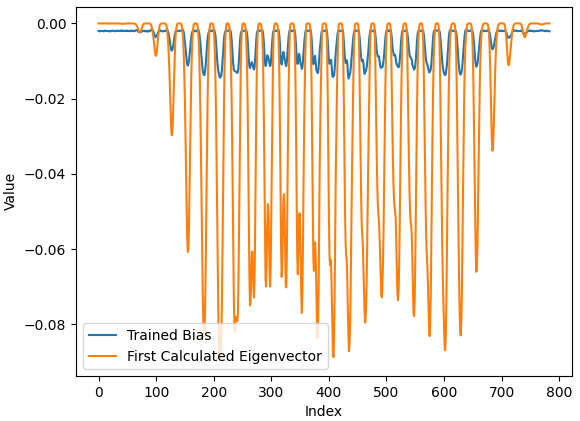}
    \caption{A plot showing the visible bias and the visible singular vectors associated to the largest singular vector.}
    \label{fig:biasVec}
\end{figure}

The link between the RBM and renormalization group coarse graining gives a rather detailed understanding of the
parameters of the network.
The key ideas are as follows
\begin{itemize}
\item[1.] Visible singular vectors are given by the eigenvectors associated to the large eigenvalues of the data
covariance matrix.
\item[2.] Hidden singular vectors are given by dropping the large Fourier modes of the visible singular vectors.
One takes the Fourier transform of the visible singular vector using the Fourier transform defined by the lattice associated
to the input data (a $28\times 28$ lattice for MNIST), drops the higher Fourier modes and performs the inverse Fourier
transform defined by the lattice associated with the output data (a $14\times 14$ lattice for our RBM).
\item[3.] The singular values tell us whether modes are relevant, marginal or irrelevant.
To estimate the singlar value associated to a visible singular vector, we perform a block spin coarse graining of the
singular vector and then use the norm of this vector as an estimate for the singular value.
\item[4.] The biases tell us what visible and hidden singular vectors are favored by the RBM probability distribution.
These should of course be related to the largest visible and hidden singular vectors. For numerical evidence supporting this identification, see Figure  \ref{fig:biasVec}.
We take the visible (hidden) biases to be equal to the largest singular value, computed in 3. above, times the corresponding
visible (hidden) singular vector. 
\end{itemize}

\begin{table}[ht]
\caption{Results of the RBM and RGM on the MNIST handwriting dataset. The RBM is trained for 8 epochs and the RGM has been trained for 2 epochs.}
\centering
\begin{tabular}{|c|c|c|c|}
\hline
Original Image & RBM & RGM & Trained RGM \\ \hline &&&\\

\includegraphics[scale=0.1]{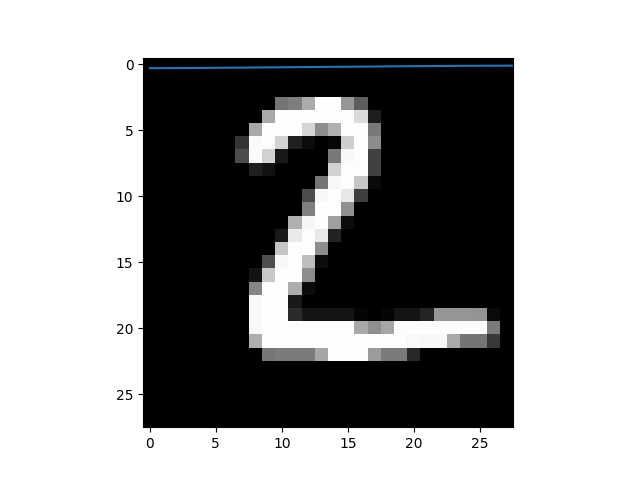}&\includegraphics[scale=0.1]{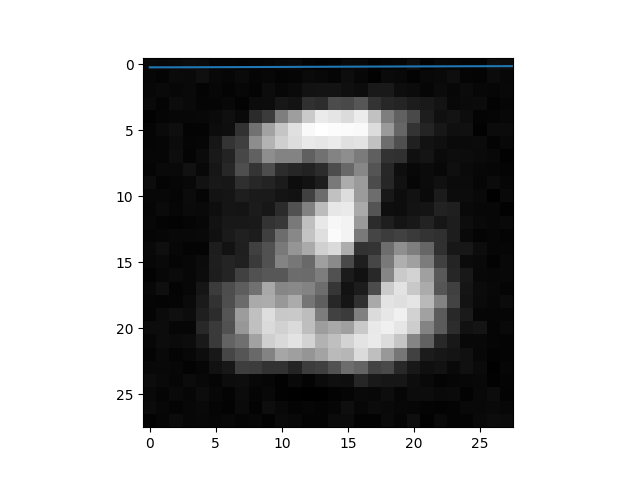}&\includegraphics[scale=0.1]{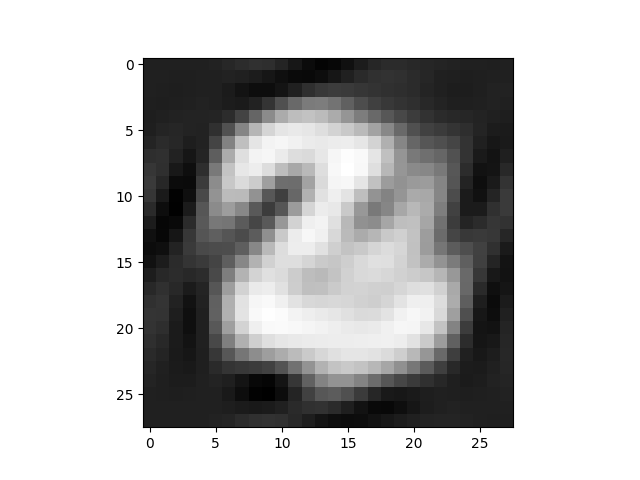}&\includegraphics[scale=0.1]{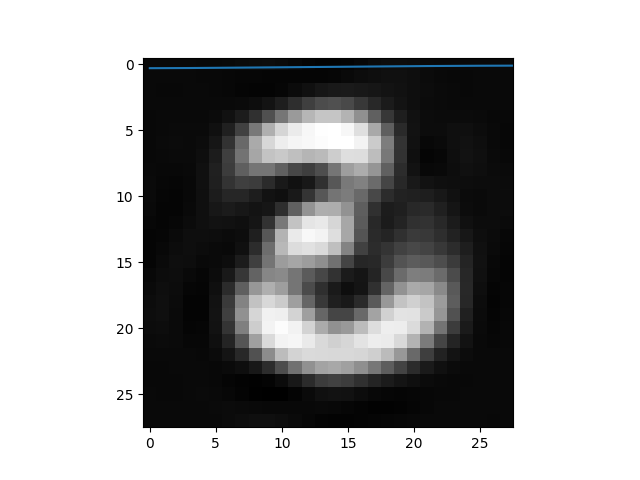}\\ 

\includegraphics[scale=0.1]{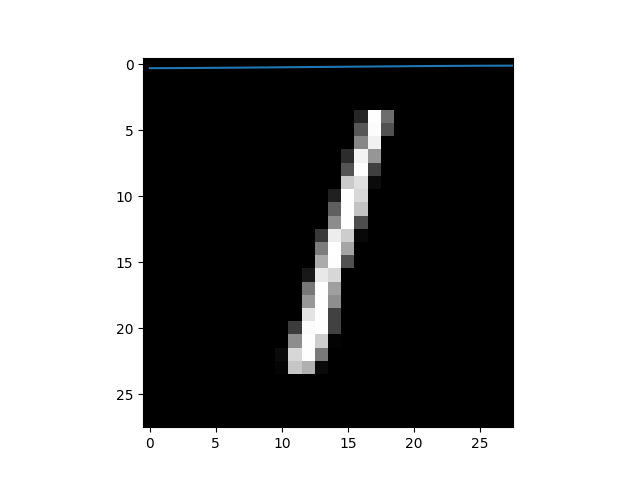}&\includegraphics[scale=0.1]{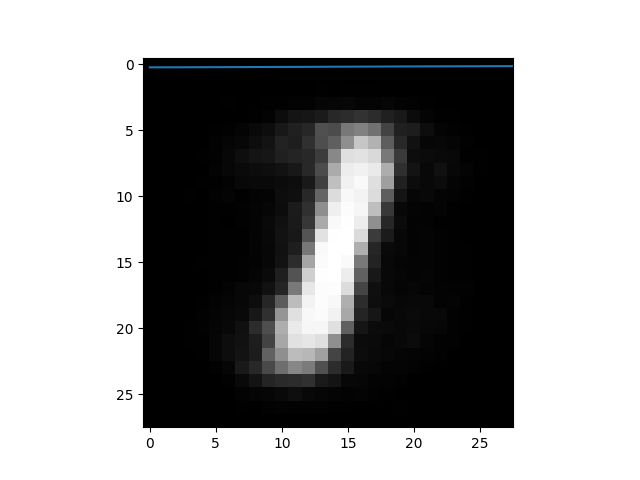}&\includegraphics[scale=0.1]{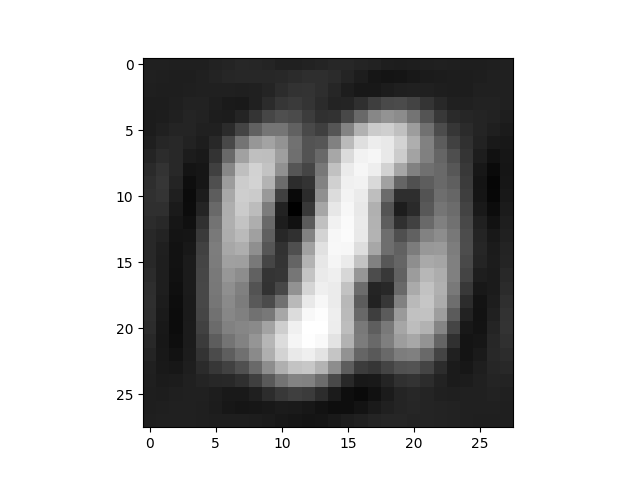}&\includegraphics[scale=0.1]{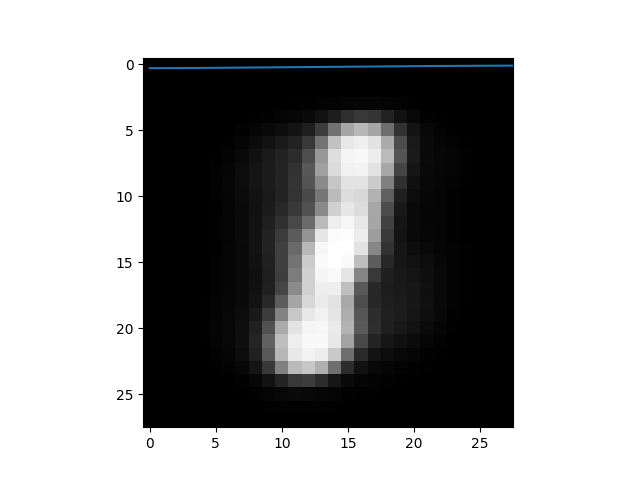} \\

\includegraphics[scale=0.1]{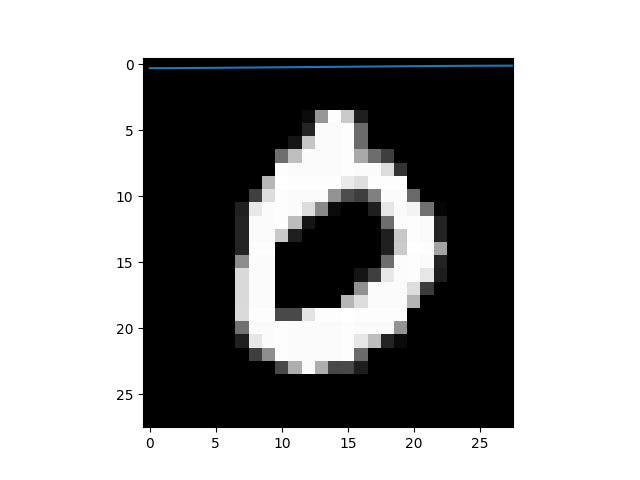}&\includegraphics[scale=0.1]{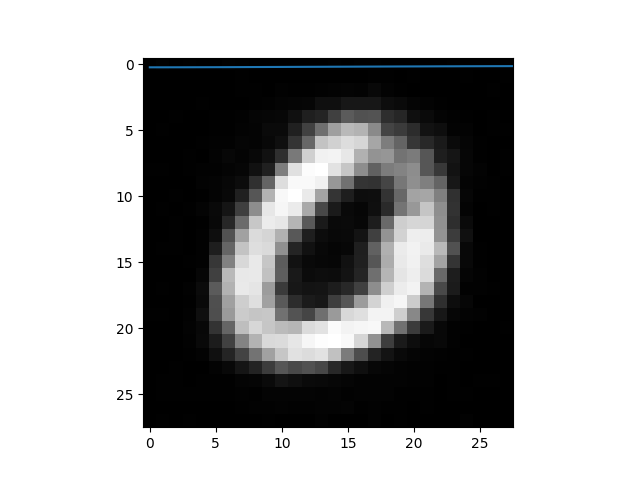}&\includegraphics[scale=0.1]{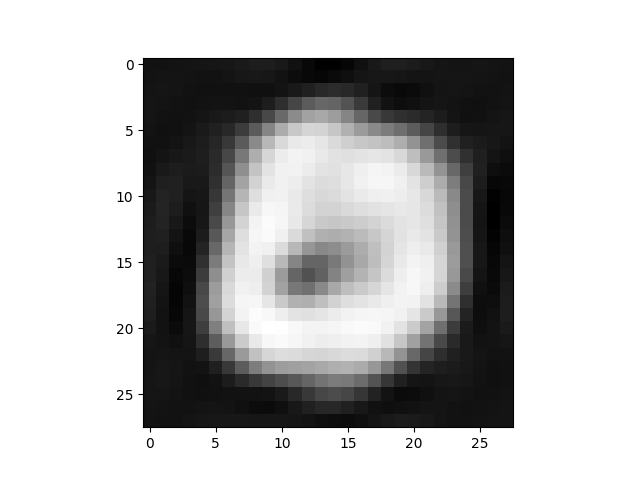}&\includegraphics[scale=0.1]{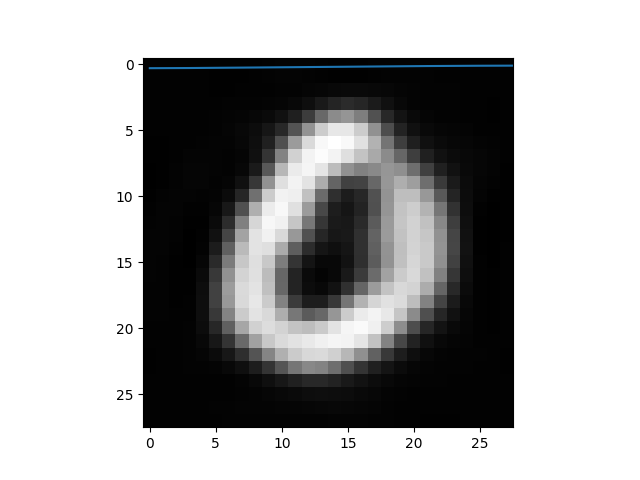} \\

\includegraphics[scale=0.1]{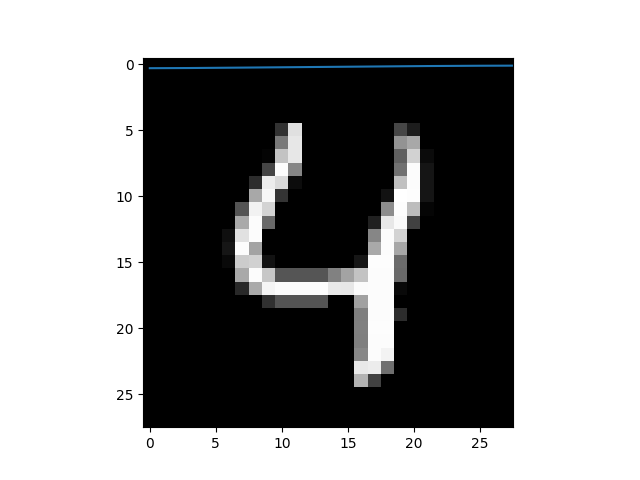}&\includegraphics[scale=0.1]{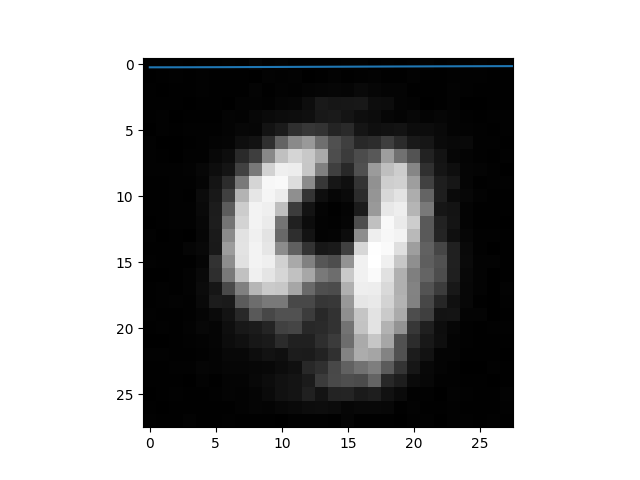}&\includegraphics[scale=0.1]{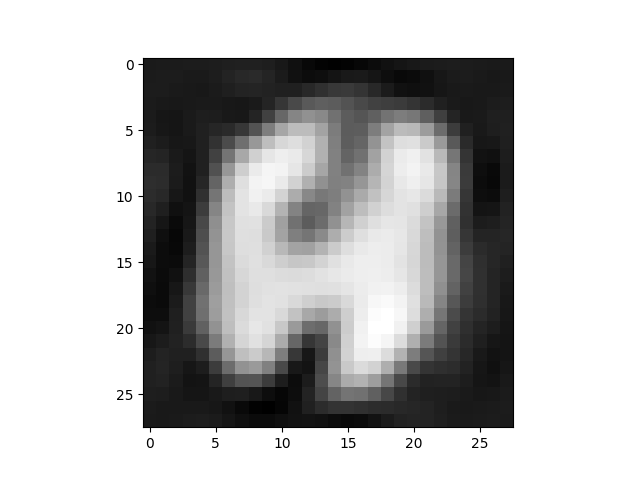}&\includegraphics[scale=0.1]{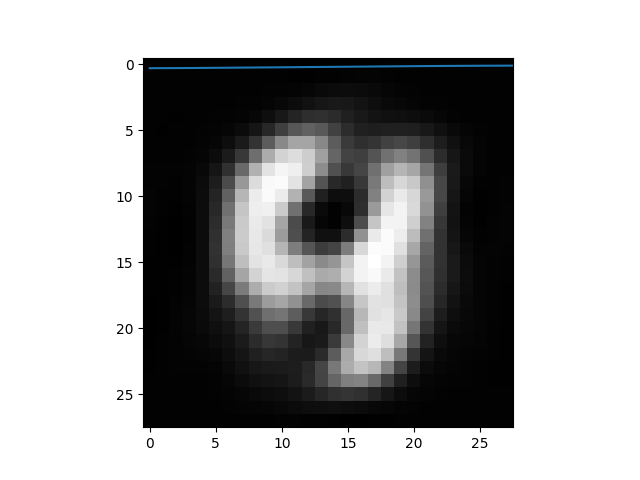} \\\hline

\end{tabular}
\end{table}

These observations provide an algorithm to choose the parameters of the RBM using only the training data set.
We call the resulting network an RGM for renormalization group machine.
For a more detailed description of the RGM see Appendix \ref{RGM}.

Table I shows a comparison of the RGM and the RBM.
The RGM is not quite correct, although the images it produces are similar to images that are produced quite late in the 
training process.
This suggests that the algorithmically computed weights are not far from the weights that are obtained from training.
In this case, the RGM may well provide a good initial condition for training.
The MNIST data set typically trains very quickly so it is not the best to use to test by how much training times are reduced.
Nevertheless, we see that training from the RGM initial condition for 2 epochs gives roughly the same performance as an
RBM trained using 8 epochs.

\subsection{Flowers}\label{flowers}

\begin{figure}[h!]
    \centering
    \includegraphics[width=0.5\textwidth]{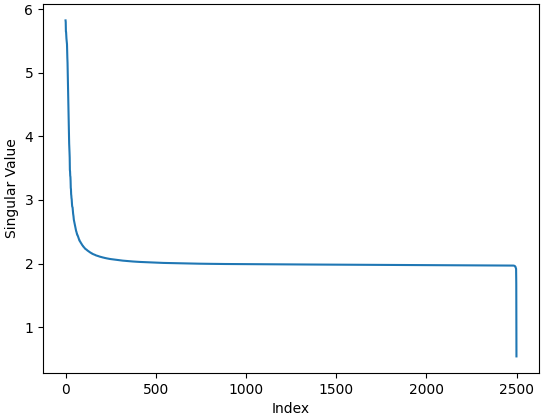}
    \caption{The singular values obtained from a singular value decomposition of the trained weight matrix of an RBM
                 trained using the TensforFlow flowers data set}
    \label{fig:flowerSV}
\end{figure}

In this section we have chosen to train an RBM on a flower data set, chosen specifically because a single layer RBM
seems to have difficulty converging to a good set of weights when trained on this data.
The TensorFlow Flowers data set \cite{tfflowers} contains $3670$ colour images 
divided into five classes which are often used for supervised learning.
These images are less correlated to each other than the images in the MNIST data set, with complicated shapes appearing 
in some images, such as humans and teapots.
We use the complete set of images to train a single layer RBM autoencoder.
The images are all rescaled to $100\times 100$, which is larger than the MNIST $28\times 28$ input, and converted 
to grayscale before they are fed into the autoencoder.
The autoencoder takes input data of size $100\times 100$ and outputs a $50\times 50$ image.

\begin{figure}[h!]
    \centering
    \includegraphics[width=0.5\textwidth]{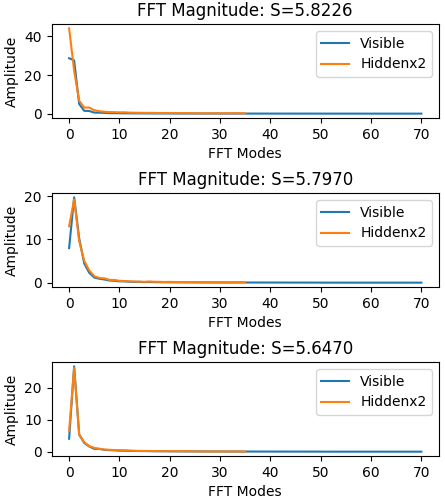}
    \caption{The radial average of the Fourier transform of the singular vectors associated to the largest singular values of the
                 RBM weight matrix trained on the TensorFlow flower data set.}
    \label{fig:flower_sing_vec}
\end{figure}

The singular value decomposition of the resulting trained weight matrix again supports the coarse graining interpretation
of the RBM.
From Figure \ref{fig:flowerSV} we again find a small number of large singular values, with a very sharp dropoff in size.
In addition, from Figure \ref{fig:flower_sing_vec} we see that the singular vectors asociated to large singular values
again have all of their support at low Fourier modes. 
The hidden singular vectors are once again given by discarding the large Fourier modes of the visible singular vectors
confirming the renormalization group link.

\begin{table}[ht]
\caption{Results of the RBM and RGM on the flower dataset. The RBM is trained for 1000 epochs and the RGM has been trained for 10 epochs.}\label{FlowerTable}
\centering
\begin{tabular}{|c|c|c|c|}
\hline
Original Image & RBM & RGM & Trained RGM \\ \hline &&&\\
\includegraphics[scale=0.1]{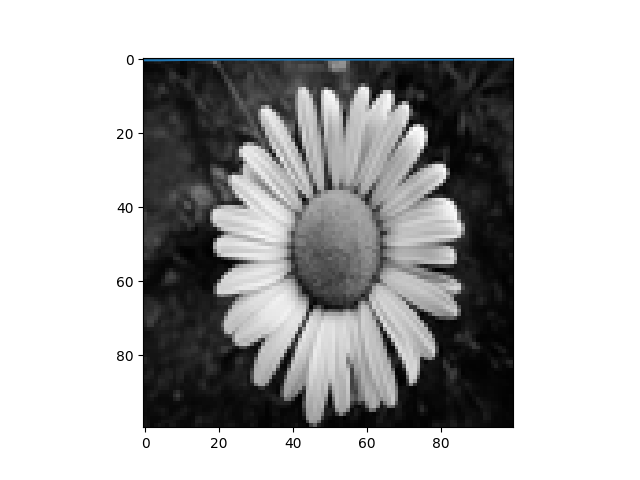}&\includegraphics[scale=0.1]{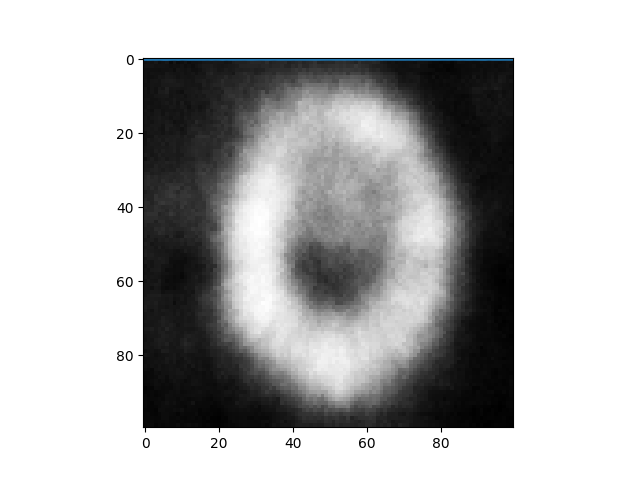}&\includegraphics[scale=0.1]{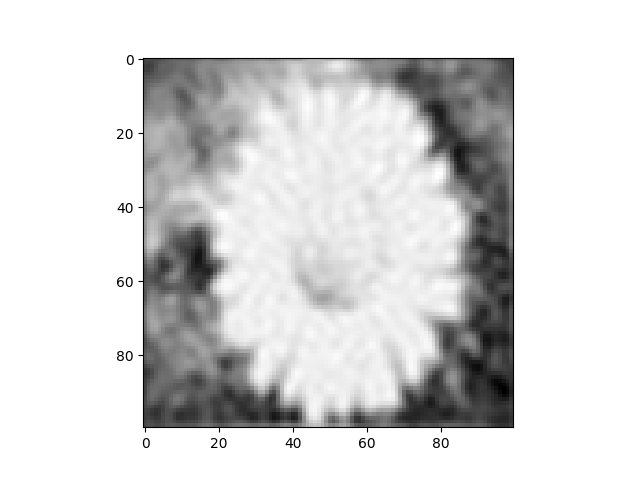}&\includegraphics[scale=0.1]{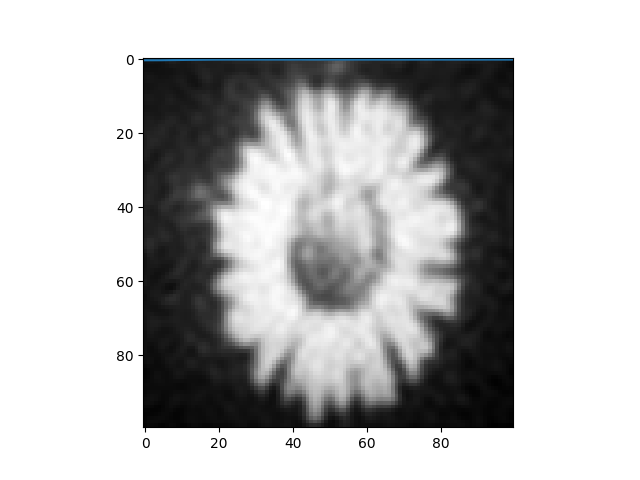}\\ 
\includegraphics[scale=0.1]{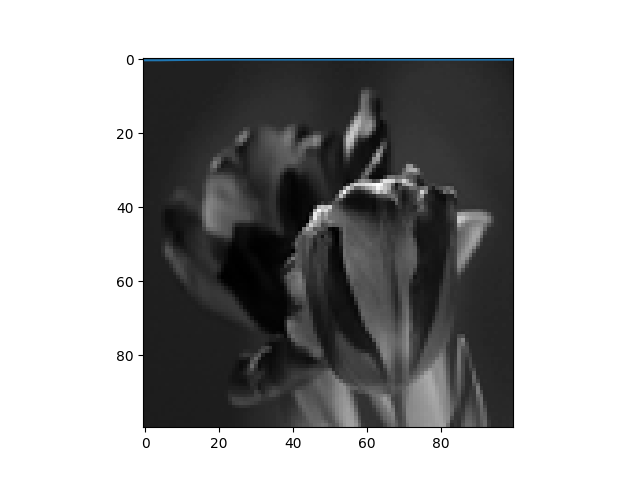}&\includegraphics[scale=0.1]{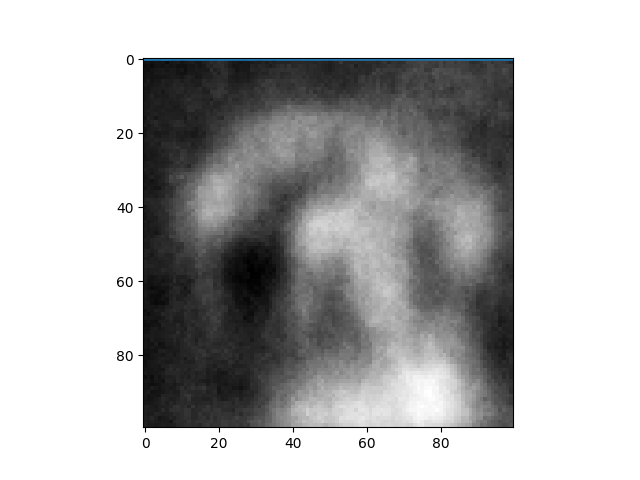}&\includegraphics[scale=0.1]{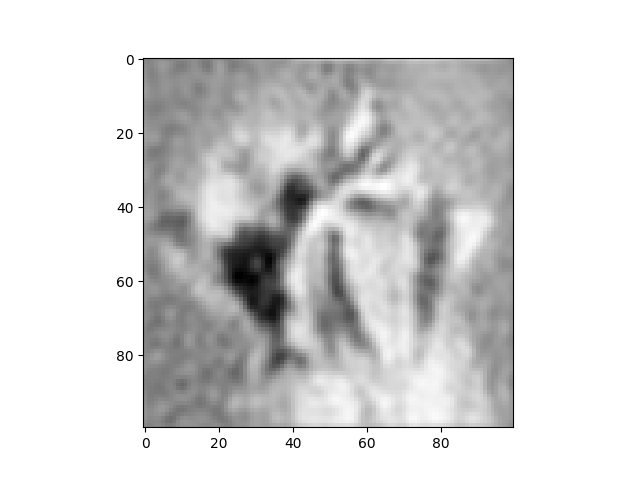}&\includegraphics[scale=0.1]{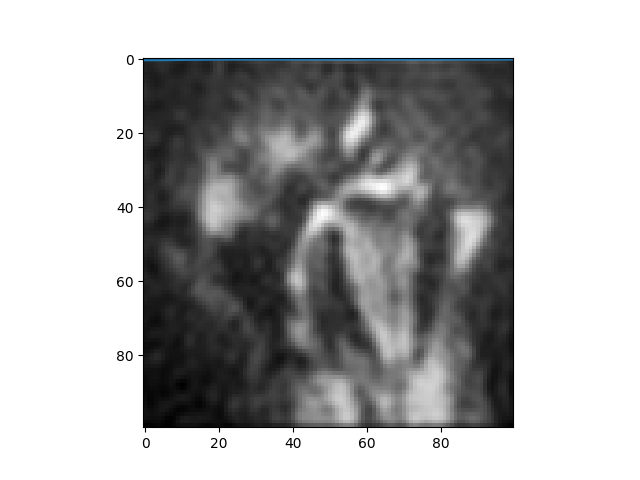} \\
\includegraphics[scale=0.1]{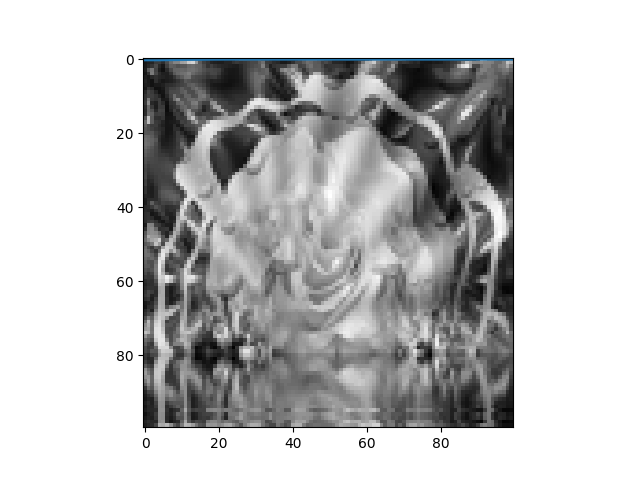}&\includegraphics[scale=0.1]{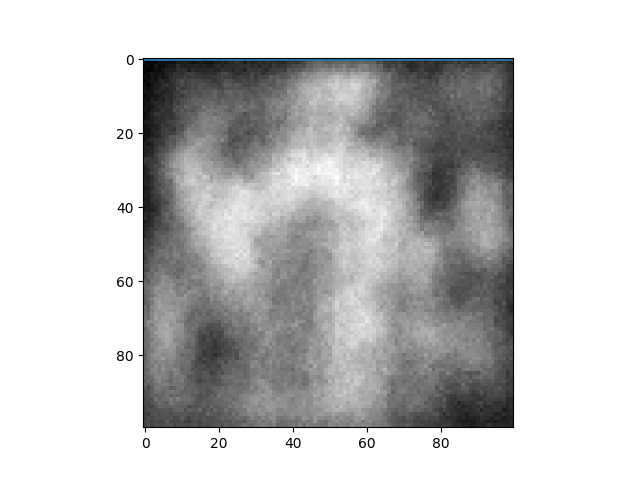}&\includegraphics[scale=0.1]{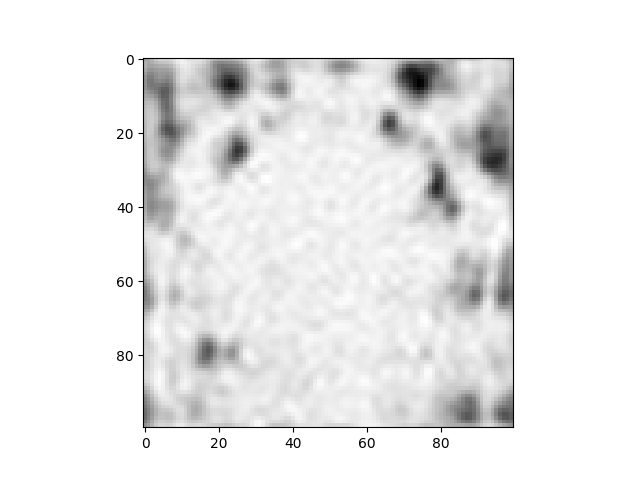}&\includegraphics[scale=0.1]{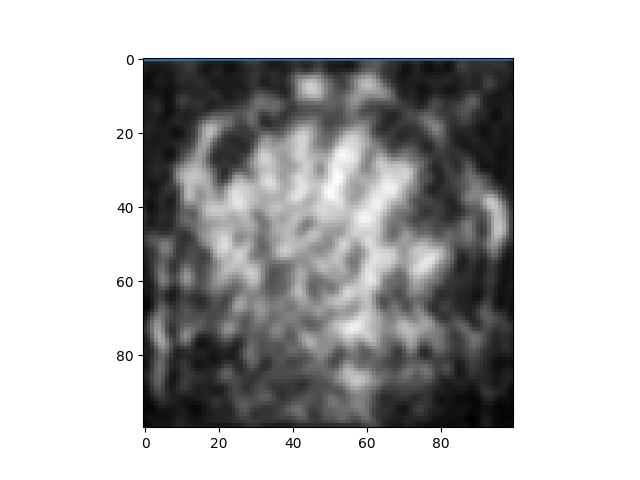} \\
\includegraphics[scale=0.1]{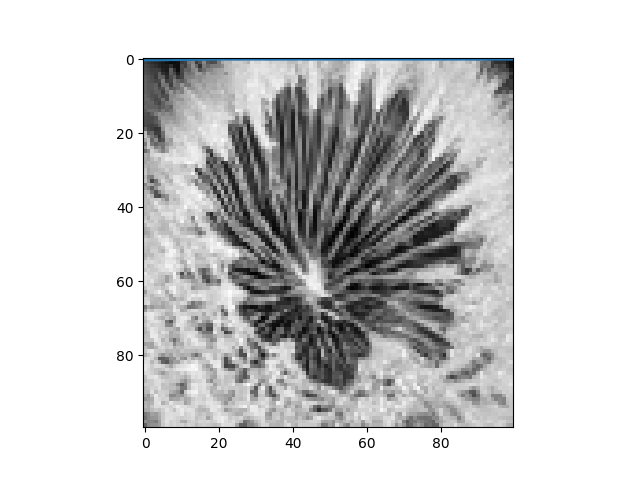}&\includegraphics[scale=0.1]{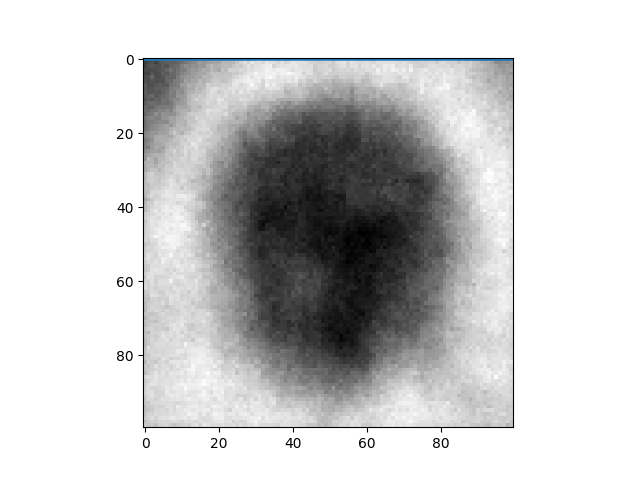}&\includegraphics[scale=0.1]{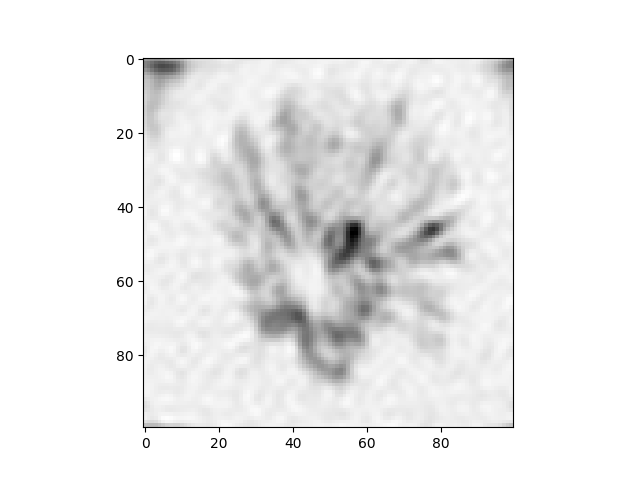}&\includegraphics[scale=0.1]{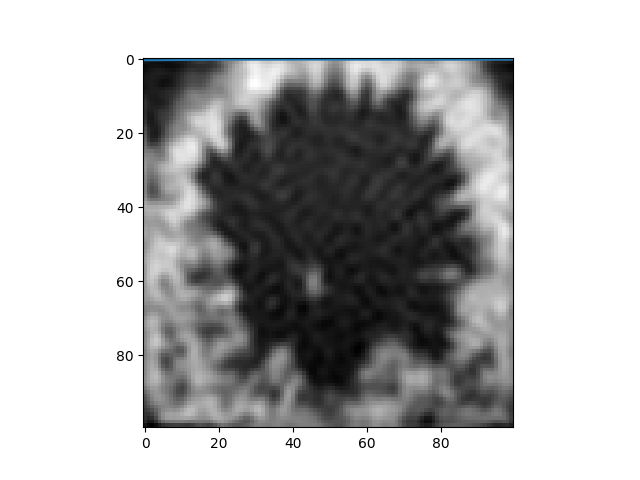} \\
\includegraphics[scale=0.1]{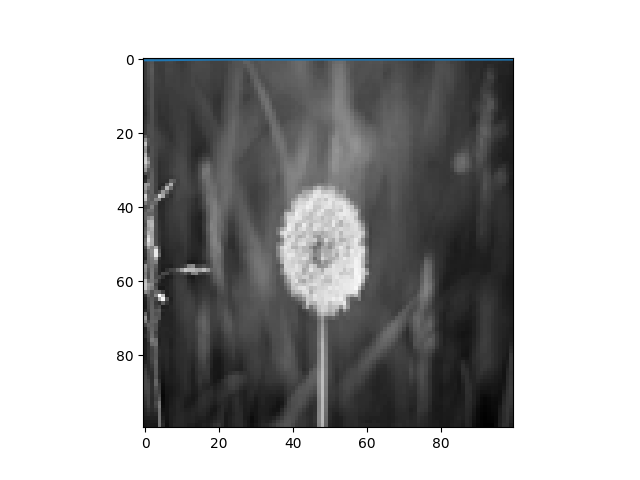}&\includegraphics[scale=0.1]{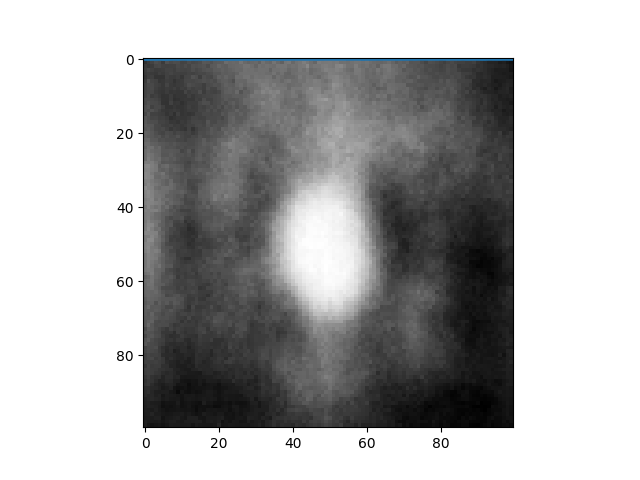}&\includegraphics[scale=0.1]{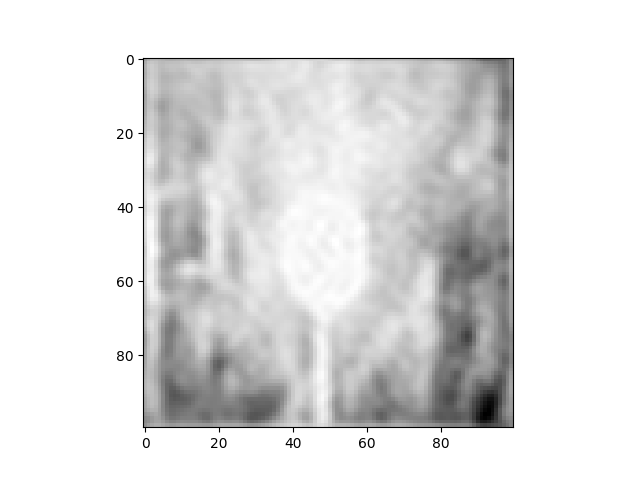}&\includegraphics[scale=0.1]{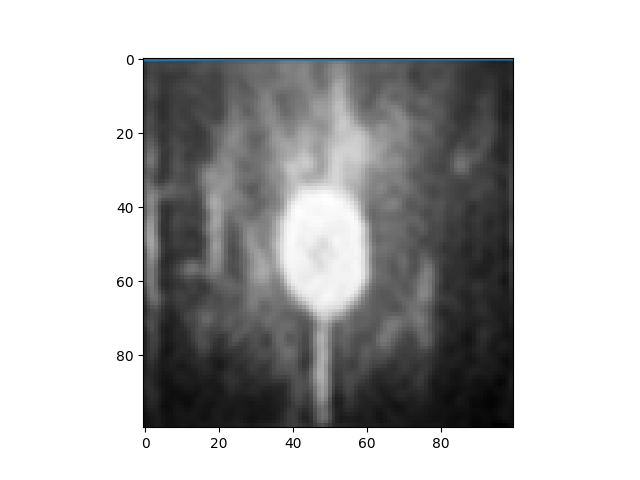}\\ 
\includegraphics[scale=0.1]{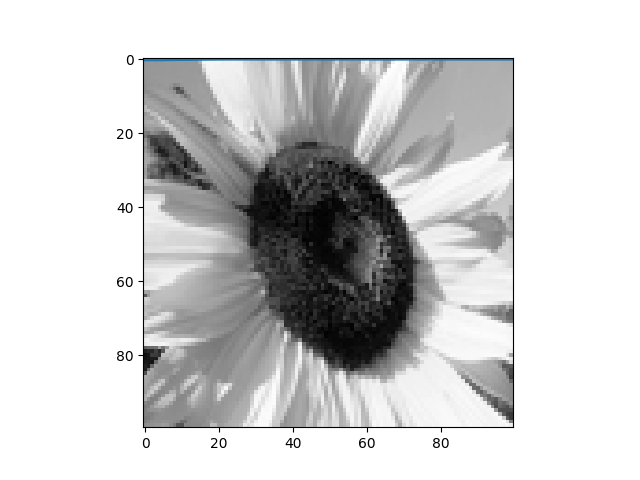}&\includegraphics[scale=0.1]{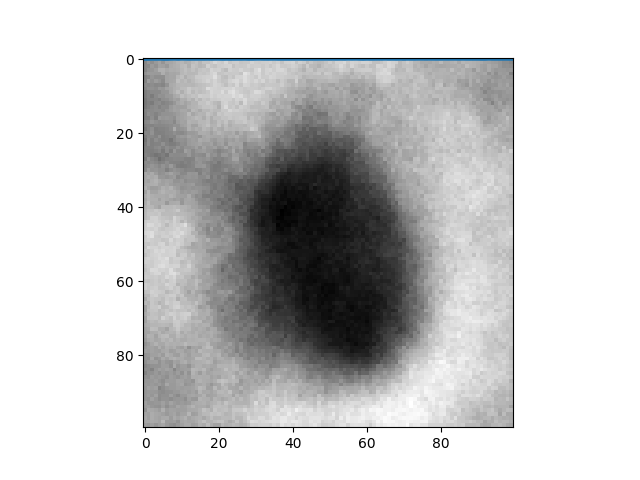}&\includegraphics[scale=0.1]{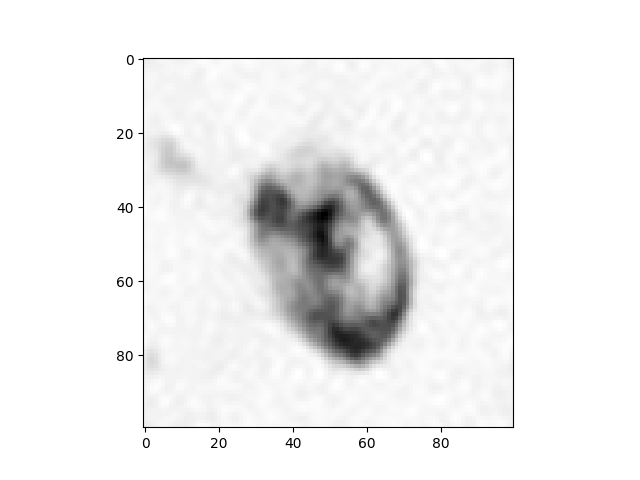}&\includegraphics[scale=0.1]{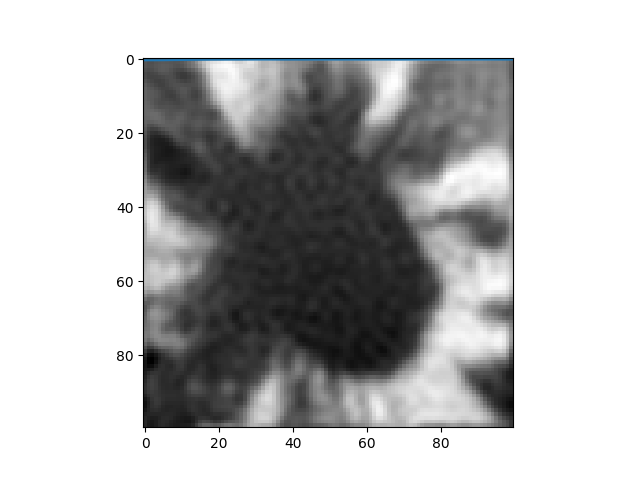}\\ 
\includegraphics[scale=0.1]{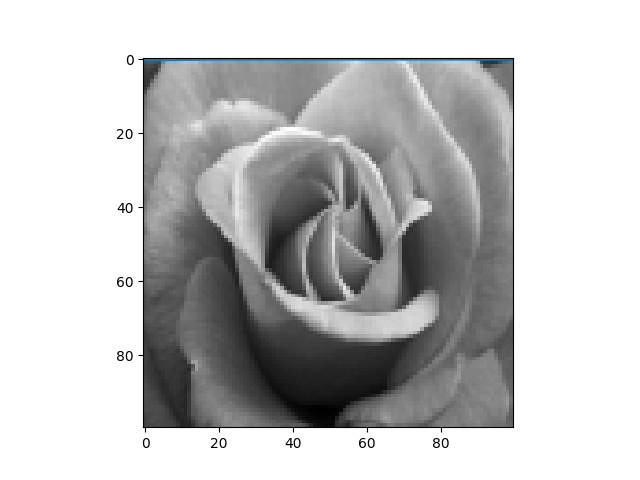}&\includegraphics[scale=0.1]{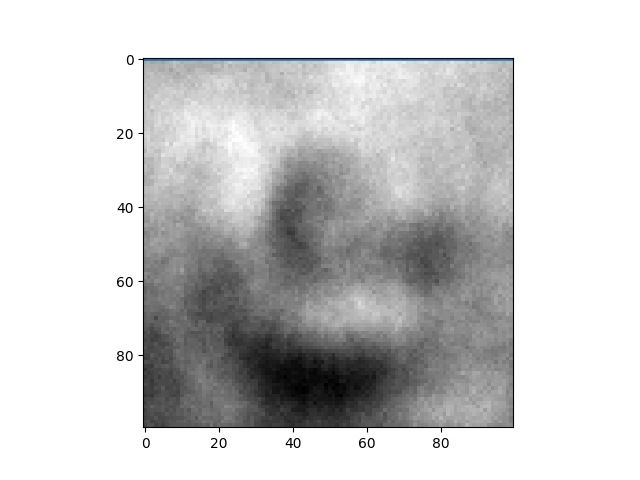}&\includegraphics[scale=0.1]{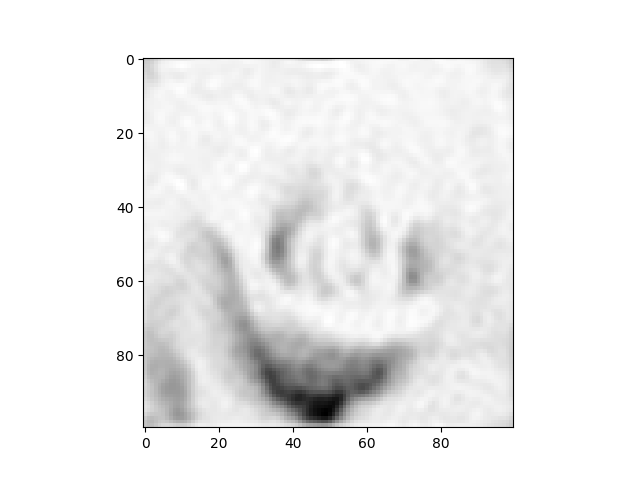}&\includegraphics[scale=0.1]{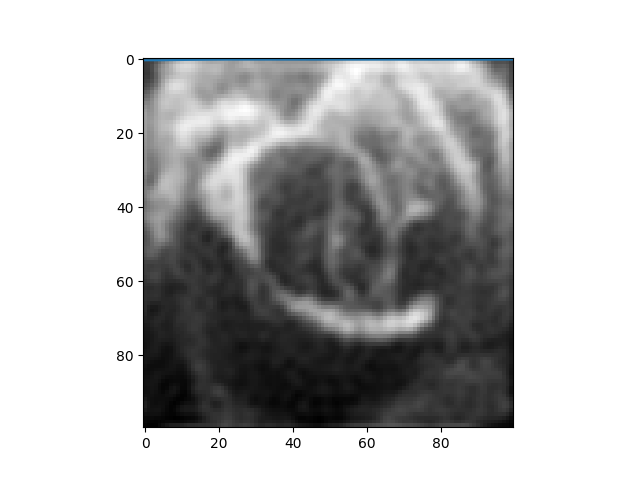} \\
\includegraphics[scale=0.1]{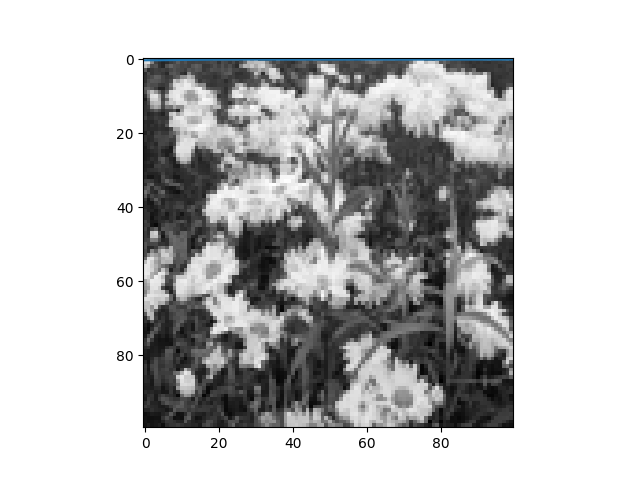}&\includegraphics[scale=0.1]{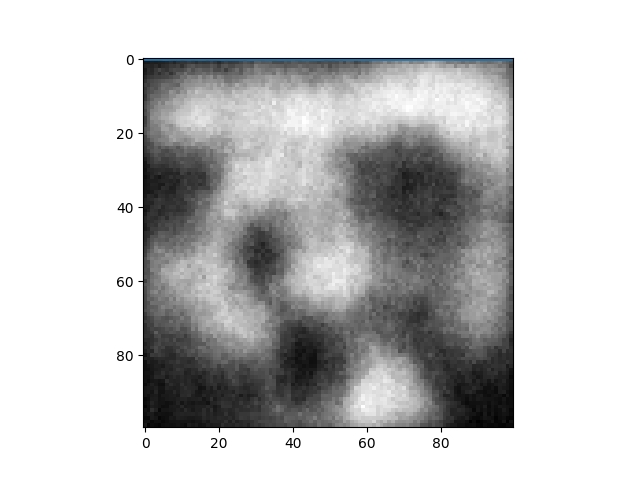}&\includegraphics[scale=0.1]{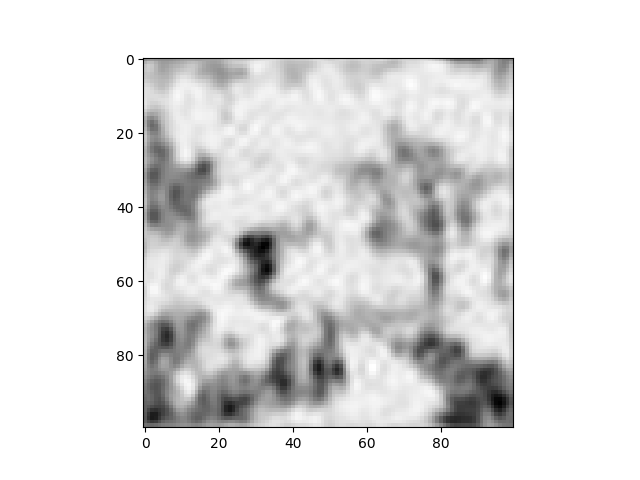}&\includegraphics[scale=0.1]{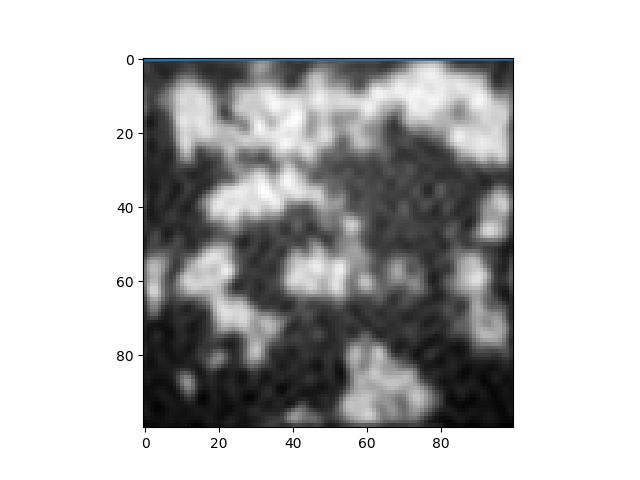} \\
\includegraphics[scale=0.1]{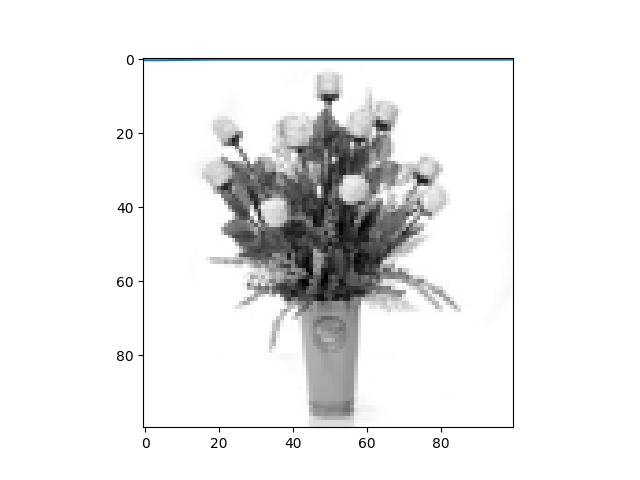}&\includegraphics[scale=0.1]{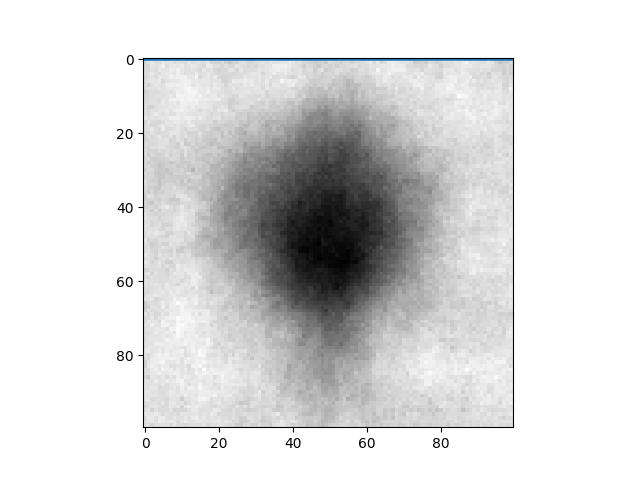}&\includegraphics[scale=0.1]{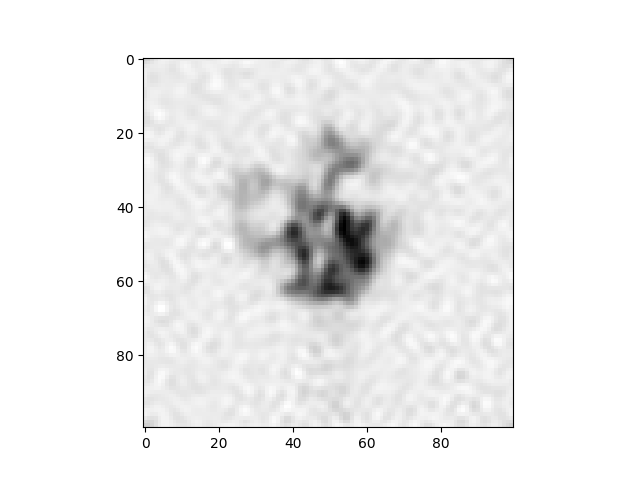}&\includegraphics[scale=0.1]{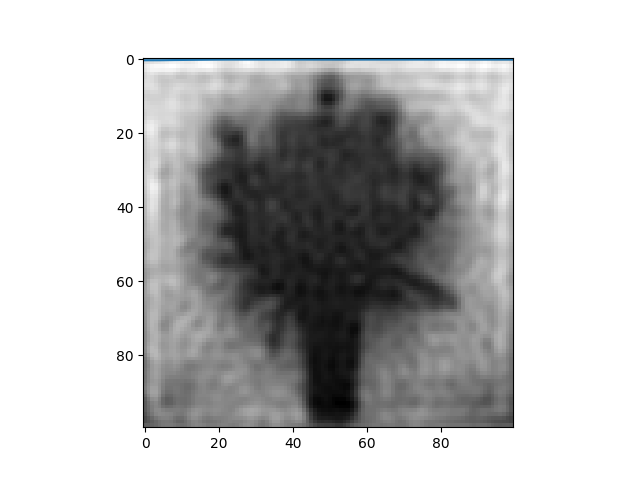} \\
\includegraphics[scale=0.1]{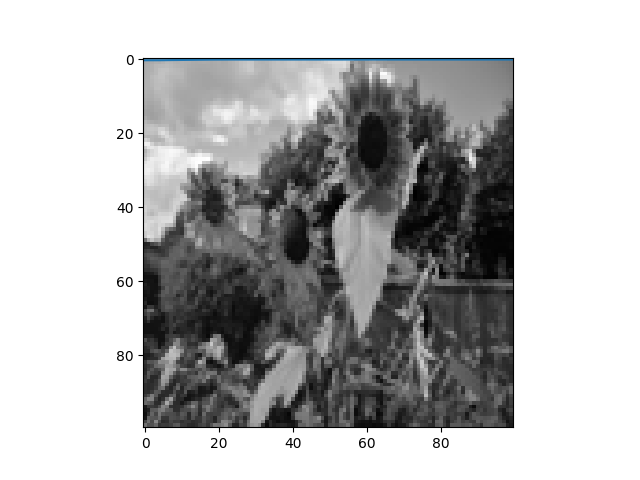}&\includegraphics[scale=0.1]{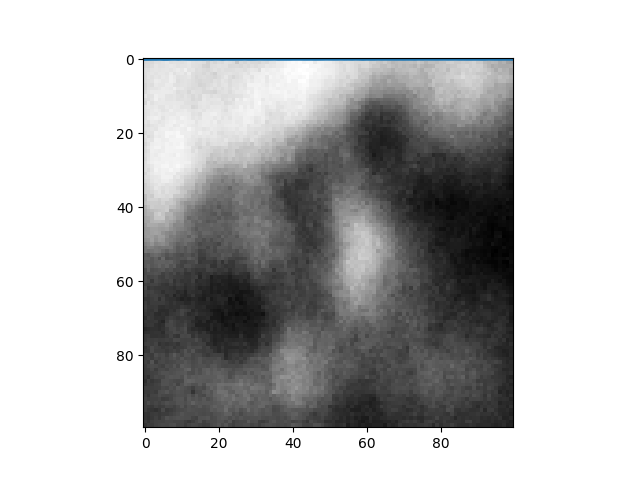}&\includegraphics[scale=0.1]{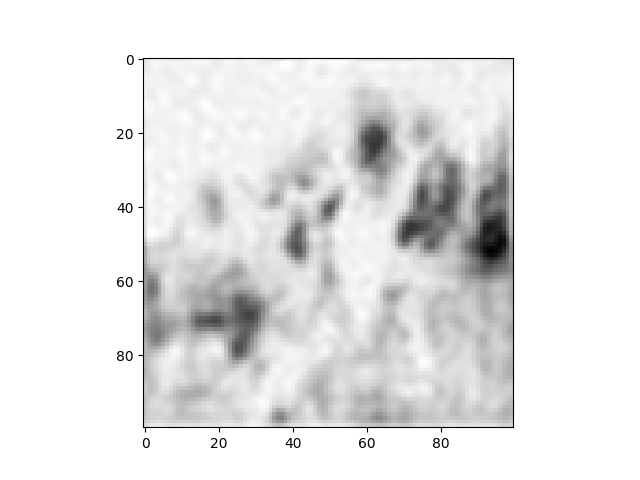}&\includegraphics[scale=0.1]{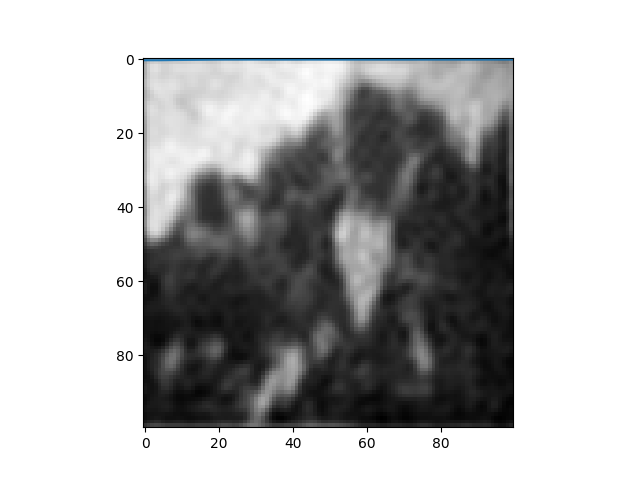} \\
\hline
\end{tabular}
\end{table}

The results of the trained RBM (trained for 1000 epochs), the RGM and the trained RGM (trained for 10 epochs)
are shown in Table \ref{FlowerTable}.
The images used to produce Table \ref{FlowerTable} did not appear in the training data set. 
The RBM used a Xavier initialization.
For many of the images, it is clear that the RGM, whose parameters are agorithmically computed from the training 
data set, outperforms the RBM trained with 1000 epochs.
This is compelling numerical support in favor of the understanding of the parameters of the RBM developed in this paper.
Further, if one now trains for 10 epochs, using the RGM as an initial condition, one obtains an autoencoder that outperforms
the trained RBM.

It is worth noting that a single layer RBM trained with the same initialization and the same data set, but with images resized to
200$\times$200 pixels, fails to converge.
The block spin coarse graining rule is a very crude approximation to the true weight matrix.
If the RBM is initialized to the block spin coarse graining rule, the RBM again converges to a valid autoencoder.
For all of the experiments we conducted, initializing to the block spin coarse graining rule give significantly better
results than the Xavier initialization, but not as good as initializing with the RGM.
This again supports the idea that the RBM is performing a coarse graining.

\subsection{Clouds}\label{clouds}

\begin{figure}[h!]
    \centering
    \includegraphics[width=0.5\textwidth]{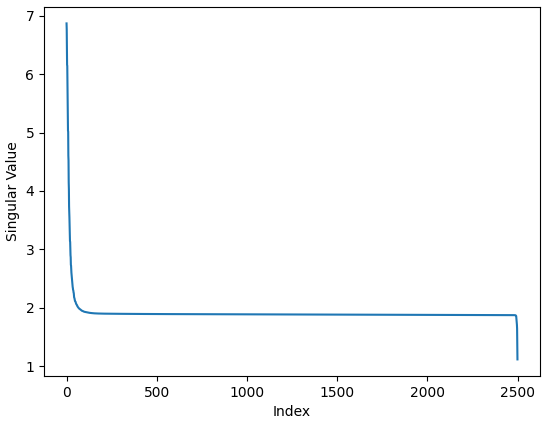}
    \caption{The singular values obtained from a singular value decomposition of the trained weight matrix of an RBM
                 trained using the SWIMSEG data set}
    \label{fig:CloudSV}
\end{figure}

\begin{figure}[h!]
    \centering
    \includegraphics[width=0.5\textwidth]{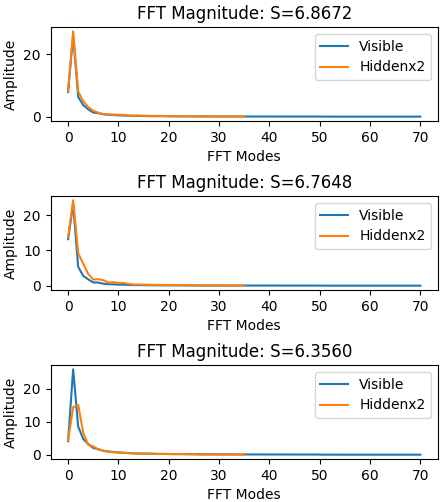}
    \caption{The radial average of the Fourier transform of the singular vectors associated to the largest singular values of the
                 RBM weight matrix trained on the SWIMSEG data set.}
    \label{fig:cloud_sing_vec}
\end{figure}

We used the Singapore Whole sky IMaging SEGmentation (SWIMSEG) data set \cite{swimseg} to train an RBM autoencoder.
Images of clouds and the sky tend to be rather homogeneous and featureless, so this is another class of images with which
to test our ideas.
SWIMSEG is made up of 1013 sky images, each $600\times 600$ pixels.
To increase the number of images, we divide each image into 36 $100\times 100$ images and convert all images to grayscale.
The autoencoder takes an input of $100\times 100$ and outputs a $50\times50$ image.
Training is initialized with a renormalization group block averaging initial condition.
From Figure \ref{fig:CloudSV} and Figure \ref{fig:cloud_sing_vec} it is again clear that the trained RBM exhibits
the hallmark features of renormalization group coarse graining.

\section{Conclusions and Discussion}\label{Conc}

A central conclusion of this study is that deep learning is performing a coarse graining.
More specifically, it is performing precisely the coarse graining of the momentum space renormalization group near
the free field fixed point.
The singular value decomposition replaces visible singular vectors with hidden singular vectors.
The hidden singular vectors are obtained by dropping the large Fourier modes of the visible singular vectors - which is 
the definition of the momentum space renormalization group near the free field fixed point.
Further, each replacement is associated with a singular value that determines when modes are relevant or irrelevant.
The large singular values, defining the relevant modes, are associated with singular vectors that have all of their
support at low Fourier modes.
These are precise mathematical statements that hold perfectly in every example we have presented.

The connection to coarse graining has implications for the generalization puzzle.
First only the relevant modes given by the large singular values need to be retained, implying a dramatic reduction in the
number of parameters, as pointed out in \cite{arora2018stronger}.
Second, relevant modes only have support at low Fourier modes, so that the coefficients of large momentum modes are
set to zero.
In addition, the coefficients appearing in the Fourier expansion of the hidden and visible singular vectors are not independent.
This implies a further dramatic decrease in the number of parameters, which has not previously been pointed out in the
literature and it is a direct consequence of the connection to the renormalization group.

The link to the renormalization group has allowed us to develop a rather detailed understanding of how the training
data set determines the parameters of the trained network.
This allows us to guess a set of network parameters given the training data set, defining what we have called a
renormalization group machine or RGM.
Although this guess is not perfect, it does give results that are close to a trained network and this guess provides a
good initial condition that can speed up training.
These results show that our study has shed light both on deep learning and on the renormalization group.
Indeed, the RGM shows how to define a renormalization group transformation for a finite data set.

Our study in this article is strongly motivated by the generalization puzzle.
Deep networks seem to have so many parameters that over fitting is inevitable.
Another facet of this puzzle is as follows: we break the data set into two disjoint subsets,
one to be used for training the network and one for testing the trained network.
There is nothing in the problem formulation that suggests that the results will be independent of how we perform the split 
into training and test data subsets.
This independence, which is present in successful deep networks, is mysterious.

An equally vexing problem is the optimization puzzle: training a deep network boils down to finding the minimum of 
a loss function, which is a function of the millions of parameters of the deep network.
We would generically expect that such a function, depending on such an enormous number of parameters, has many local
minima and that a simple learning rule would almost always fail to take us to the global minimum.
The different minima would correspond to different ways to thread a surface through the data.
Typical deep learning problems, which do use simple training rules - typically some form of gradient descent implemented
using back propagation - do manage to find great solutions and hence the puzzle.

A tacit assumption made by both puzzles is that deep learning is a form of curve fitting.
In our opinion these puzzles suggest that curve fitting is the wrong paradigm to explain deep learning.
Our study suggests an alternative point of view: deep learning is a sophisticated coarse graining which extracts
trends and meaning from enormous data sets\footnote{See 
\cite{mehta2014exact,iso2018scale,funai2018thermodynamics,lin2017does,koch2020deep,de2020short} 
where this idea was already considered.}. 
Consider an analogy which illustrates the idea.
Imagine for the sake of example, that our system is a container filled with water.
The state of the water is specified by an enormous list of positions and velocities of the water molecules, and the fine
grained description uses the masses of the water molecules, as well as detailed and complicated interaction potentials.
The coarse grained description uses a handful of parameters: things like the density of the fluid, its viscosity, the
speed of sound in the fluid and its state is specified by a much smaller list of quantities like pressure and temperature
of the water.
Assuming the water is at equilibrium, we can choose any (large enough) subregion of the water (the training data) and use 
it to determine the parameters of the coarse grained description (the trained network).
The rest of the system (test data) can then be used to test how good the description provided by the coarse grained theory is.
Within the setting of coarse graining to obtain an effective description, it is natural to ask if the analysis depends on the details
of the split into test and training data.
Assuming that an effective theory exists and since the water is at equilibrium, we are certain that our results are independent 
of the details of this split, and hence generalization is a natural question and is guaranteed.
There will obviously be fluctuations in the effective theory parameters, but these are small for large systems, 
so that to very good accuracy we get the same effective theory no matter how the training data is chosen.
Thus, within the coarse graining framework, the generalization puzzle is a natural question and it is convincingly resolved.

Let us return to the optimization puzzle.
By curve fitting logic, we expect there are many different minima, each corresponding to a different way of threading
a surface through the data.
It is inconceivable that these different surfaces give the same prediction on unseen data.
On the other hand, coarse graining logic suggests that the network is finding the subspace on which the data is located.
It is true that many of the parameters of networks trained on different training data sets will differ, because they 
will have different values for the irrelevant parameters.
The point however, is that these networks will still agree on unseen data to good accuracy, because they have both learned 
the data subspace.
This corresponds to a famous fact known as universality of the renormalization group flow: many different UV theories
flow to the same IR theory.

The above discussion suggests a simple criterion to decide if a given problem can be solved or not.
The criterion basically checks to see if the effective description constructed by coarse graining is well defined. 
One starts by selecting data samples at random, from the training data set, and computing their data covariance matrix.
Once the number of large eigenvalues of the data covariance matrix is no longer increasing as additional samples are
selected, the process is terminated and the eigenvectors associated to the large eigenvalues are recorded.
The process is then repeated, to produce a second set of eigenvectors associated to the large eigenvalues.
If no matter how many times the process is repeated, the eigenvectors that are produced span the same subspace, the 
problem has a well defined effective description and it can be solved using a network.
It would be interesting to explore the relation between this criterion and the one presented in 
\cite{arora2019fine}.

Experience with the renormalization group suggests more.
It relates a short distance, high energy theory to a long distance, low energy theory.
According to our analogy, states of the high energy theory correspond to elements of the input data set, while the network
outputs states of the coarse grained effective theory.
The renormalization group splits parameters of the high energy theory into relevant, marginal and irrelevant parameters,
according to how the coarse graining affects the value of the parameters.
Relevant parameters grow under coarse graining, irrelevant parameters go to zero and marginal parameters maintain their size.
The renormalization group explains why there are a handful of relevant or marginal parameters, but many irrelevant
parameters.
Changing the input data, for example by adding noise, changes the state of the high energy theory and hence it corresponds
to changing parameters.
Almost all changes of the input data correspond to changes in irrelevant couplings which will not change the low energy state,
i.e. the output of the deep network.
In this way the coarse graining framework predicts noise rejection of deep networks, which is established in practice.

There are a number of interesting directions that can be pursued.
In what follows we list some of them.
\begin{itemize}
\item In this article we have focused on unsupervised learning. 
It would be fascinating to explore supervised learning in detail, in the coarse graining framework.
If this framework is applicable, a key question is how the labels of the data are used to determine 
the relevant modes of the coarse grained description.
There are also a number of simple tests that can be performed to test whether or not the coarse grained description is
the correct one. For example, does one still see a large number of small singular values for the weights of fully connected
layers? Further, if the weight matrix is initialized correctly, does one still find that the singular vectors associated to large
singular values have their support on low Fourier modes? Are the hidden singular vectors still obtained from the visible 
singular vectors by discarding large Fourier modes? 
\item Our study considers learning data sets composed of images. 
The pixels are organized in a two dimensional lattice.
This is an important ingredient: for the Fourier transform the data needs to be organized in a sensible two dimensional lattice.
For more general applications there might be no obvious way to define the Fourier transform.
This is an interesting direction that deserves to be explored.
\item The role of depth of the network can be studied in more detail.
For stacked RBMs, where the deep network is trained
layer-wise, we have seen that each layer is performing a renormalization group transformation.
Thus, a stacked network is performing a renormalization group flow.
Renormalization group flows can have a rich behavior, with the relevance and irrelevance 
of modes changing as the flow proceeds.
It would be interesting to see if similar phenomena appear in a deep network.
\item Our guess for the network parameters, in the form of the RGM, is close to what is produced by training.
By improving our understanding of what determines the singular values of the trained weight matrix, 
it may be possible to further improve this guess.
Giving an algorithm for the network parameters in terms of the training data, which is as good as parameters learned from
training maybe too ambitious.
What is however clear is that progress along this direction may lead to initial conditions that significantly reduce training times. 
\end{itemize}
This is a rather incomplete list of questions.
The coarse graining framework suggests many more.
We look forwards to exploring these in the future.

$$ $$

\section*{Acknowledgement}
This work is supported by the Science and Technology Program of Guangzhou (No. 2019050001 and
No. 2020A1515010388), by the National Natural Science Foundation of China under Grant No. 12022512 and No. 12035007,
by a Simons Foundation Grant Award ID 509116 and by the South African Research Chairs initiative 
of the Department of Science and Technology and the National Research Foundation.
We are grateful for useful discussions and encouragement to Byron Rudman, Dimitris Giataganas, Jeff Murugan, 
Junggi Yoon,
Lin Hong de Mello Koch, Nicholas Kastanos, Roger de Mello Koch and Sanjaye Ramgoolam. 

\appendix

\section{Restricted Boltzmann Machines}\label{RBM}

Restricted Boltzmann Machines (RBMs) consist of two layers of nodes - a visible layer (input nodes) and a hidden layer 
(output nodes). Connections are allowed between every visible and hidden node but no connections are allowed between 
nodes within the same layer. 
The energy function for the RBM is given by 
\begin{equation}
    E=-\sum_{i=1}^{N_v}\sum_{a=1}^{N_h}v_iW_{ia}h_a -\sum_{i=1}^{N_v}v_ib_i^{(v)}-\sum_{a=1}^{N_h}h_ab_a^{(h)}
    \label{eq:rbm_energy}
\end{equation}
where $W_{ia}$ is the weight matrix which determines strengths between the ith visible, $v_i$, and ath hidden node, $h_a$, $b_i^{(v)}$ is the bias of the ith visible node and $b_a^{(h)}$ the bias of the ath hidden node.
The visible and hidden nodes take values of $\pm 1$.

The probability of a given visible and hidden vector is 
\begin{equation}
    p(v,h)=\frac{1}{Z}e^{-E}
\end{equation}
where the normalization $Z$ is the partition function 
\begin{equation}
    Z=\sum_{\{v,h\}}e^{-E}
\end{equation}
The marginal distribution of a visible or hidden vector is given by summing over the space of hidden and visible vectors respectively
\begin{equation}
    p(v)=\frac{1}{Z}\sum_{\{h\}}e^{-E}
\end{equation}
\begin{equation}
    p(h)=\frac{1}{Z}\sum_{\{v\}}e^{-E}
\end{equation}

The network is trained using a technique called contrastive divergence which approximates minimizing the Kullback-Leibler divergence. 
The goal of training is to match the distribution of the model to the distribution of the data.
The Kullback-Leibler divergence measures how similar two distributions are and is given by
\begin{equation}
    \begin{split}
        D_{KL}(q||p)&=\sum_{\{v\}}q(v) (\log(q(v))-log(p(v)))\\
        &=\sum_{\{v\}}q(v)\log\left( \frac{q(v)}{p(v)} \right)
\end{split}
\end{equation}
where $q(v)$ defines the data distribution and $p(v)$ denotes the model distribution we determine during training.
The Kullback-Leibler divergence is always not negative and only vanishes if the two distributions are equal.
Training tunes the model parameters $W_{ia}$, $b_i^{(v)}$ and $b_a^{(h)}$ to minimize $D_{KL}$.
We tune parameters by taking derivatives of $D_{KL}$ with respect to each parameter
\begin{equation}
    \begin{split}
        \frac{\partial D_{KL}}{\partial W_{ia}} = \corr{v_ih_a}_{data}-\corr{v_ih_a}_{model}
    \end{split}
\end{equation}
\begin{equation}
    \begin{split}
        \frac{\partial D_{KL}}{\partial b_{i}^{(v)}} = \corr{v_i}_{data}-\corr{v_i}_{model}
    \end{split}
\end{equation}
\begin{equation}
    \begin{split}
        \frac{\partial D_{KL}}{\partial b_{a}^{(h)}} = \corr{h_a}_{data}-\corr{h_a}_{model}
    \end{split}
\end{equation}
The right hand side of these expressions can now be evaluated.
It is impractical to perform the sum over the state space of visible and hidden vectors, because the size of each state space is
enormous. 
It is for this reason that we employ contrastive divergence (CD) to approximate the KL divergence. 
CD allows us to sum over the space of input vectors present in the training set rather than the entire state space of visible vectors.
To approximate the state space of hidden vectors we sample a set of hidden vectors, generated using the input 
(visible) vectors.

Given a set of visible vectors we can sample the hidden vectors using the probability 
\begin{equation}
    p(h_a=1|v) = \frac{1}{2}\left(1+\tanh\left(\sum_iW_{ia}v_i+b_a^{(h)}\right)\right)
\label{eq:p_of_h}
\end{equation}
Given a set of hidden vectors we can sample a set of visible vectors using the probability
\begin{equation}
    p(v_i=1|h) = \frac{1}{2}\left(1+\tanh\left(\sum_aW_{ia}h_a+b_i^{(v)}\right)\right)
\label{eq:p_of_v}
\end{equation}

To calculate expectation values over the data we use the input training set of visible vectors denoted $\hat{v}$.
To generate the set of hidden vectors used in the data averages, $\hat{h}$, we sample hidden vectors using equation \eqref{eq:p_of_h} and $\hat{v}$.

We then sample the set of visible vectors which are used in the model averages, $\tilde{v}$, using $\hat{h}$ and equation \eqref{eq:p_of_v}. 
Using $\tilde{v}$ and equation \eqref{eq:p_of_h} we sample hidden nodes to be used in the model expectations, $\tilde{h}$.

Contrastive divergence now amounts to the following sequence of approximations
\begin{equation}
    \corr{v_ih_a}_{data}=\frac{1}{N_s}\sum_{A=1}^{N_s}\hat{v}_i^{(A)}\hat{h}_a^{(A)}
\end{equation}
\begin{equation}
    \corr{v_ih_a}_{model}=\frac{1}{N_s}\sum_{A=1}^{N_s}\tilde{v}_i^{(A)}\tilde{h}_a^{(A)}
\end{equation}
\begin{equation}
    \corr{v_i}_{data}=\frac{1}{N_s}\sum_{A=1}^{N_s}\hat{v}_i^{(A)}
\end{equation}
\begin{equation}
    \corr{v_i}_{model}=\frac{1}{N_s}\sum_{A=1}^{N_s}\tilde{v}_i^{(A)}
\end{equation}
\begin{equation}
    \corr{h_a}_{data}=\frac{1}{N_s}\sum_{A=1}^{N_s}\hat{h}_a^{(A)}
\end{equation}
\begin{equation}
    \corr{h_a}_{model}=\frac{1}{N_s}\sum_{A=1}^{N_s}\tilde{h}_a^{(A)}
\end{equation}

\section{RG}\label{RG}

In this Appendix we will give a brief review of the renormalization group.
The goal of the renormalization group is to derive a long distance effective description, given some underlying microscopic 
dynamics.
This typically happens by coarse graining the short distance degrees of freedom. 
The name ``the renormalization group'' is a bad one since the mathematical 
structure of the renormalization group is not that of a group,
renormalization of quantum field theory is not an essential ingredient\footnote{One of the earliest applications of these
methods is to quantum field theory, which is the origin of the name.} and the method is not unique but rather it is a set
of ideas that must adapted to the nature of the problem at hand.
In the next section we will review the method as it applies to Euclidean quantum field theory.
We have chosen to review this material because it is the basic mechanism at work in fully connected layers and, further,
it is a good example of how the renormalization group explains why so many parameters of the short distance theory 
simply have no effect on the effective description.
We have interpreted this as the basic mechanism behind understanding why deep networks generalize.
We go on to describe the block spin transformation implementation of the renormalization group coarse graining
as it is directly relevant to the discussion of the Ising model in Section \ref{ising}. 

\subsection{Momentum Space RG}\label{momSpaceRG}

Momentum space RG is a tool used routinely in quantum field theory and statistical mechanics \cite{wilson1974renormalization}. 
The method coarse grains by first organizing the theory according to length scales and then averaging over the short distance
degrees of freedom, which gives an effective theory for the long distance degrees of freedom. 

To illustrate the procedure, we will consider a quantum field theory, describing a quantum field $\phi$.
Observables ${\cal O}$ are functions (usually polynomials) of the field and its derivatives.
Examples of observables are the energy or momentum of the field, or correlation functions of the field.
To calculate the expected value $\langle {\cal O}\rangle$ of observable ${\cal O}$, integrate (i.e. average) over all 
possible field configurations
\bea
\langle {\cal O}\rangle =\int [d\phi] e^{-S} {\cal O}
\eea
The factor $e^{-S}$, which defines a probability measure on the space of fields, depends on the theory considered.
$S$ is called the action of the theory. 
The action is a polynomial in the field and its derivatives.
The coefficients appearing in this polynomial are the parameters of the theory, things like couplings and masses.
These are the parameters of the short distance theory.
A theory is defined by specifying the action $S$.

To coarse grain, express the position space field in terms of momentum space components
\bea
\phi (x)=\int dk e^{ik\cdot x}\phi (k)
\eea
$e^{ik\cdot x}$ oscillates in position space with wavelength ${2\pi\over k}$.
High momentum (big $k$) components have small wavelengths and encode small distance structure. 
Low momentum components have huge wavelengths and describe large distance structure.
Declare there is a smallest possible structure, implemented by cutting off the momentum modes
at a large momentum $\Lambda$ as follows
\bea
[d\phi ]=\prod_{k<\Lambda} d\phi (k)
\eea
The renormalization group breaks the integration measure into high and low momentum components
$[d\phi ]=[d\phi_<][d\phi_>]$ where
\bea
[d\phi_<]&=&\prod_{k< (1-\epsilon)\Lambda} d\phi (k)\cr
[d\phi_>]&=&\prod_{(1-\epsilon)\Lambda<k<\Lambda} d\phi (k)\label{ssplit}
\eea
By assumption, we consider observables that depend only on large scale structure of the theory, i.e.
observables that depend only on $\phi_{<}$.
In this case, when computing the expected value of ${\cal O}$ we can pull ${\cal O}$ out of the integral over
$\phi_{>}$ and integrate over the high momentum components
\bea
\langle {\cal O}(\phi_{<})\rangle
&=&\int [d\phi_<] \int [d\phi_>]\,\,  e^{-S} {\cal O}(\phi_{<})\cr
&=&\int [d\phi_<]\,\, e^{-S_{\rm eff}} {\cal O}(\phi_{<})\label{outershell}
\eea
This procedure of splitting momentum components into two sets and integrating over the large momenta
defines a new action $S_{\rm eff}$.
Repeating the procedure many times defines the RG flow under which $S_{\rm eff}$ changes continuously.
After the flow, one is left with an integral over the very long wavelength modes.
This completes the coarse graining: we have a new theory defined by $S_{\rm eff}$.
The new theory uses only long wavelength components of the field and correctly reproduces the expected
value of any observable depending only on long wavelength components.

Values of the parameters of the theory, which appear in $S_{\rm eff}$ change under this transformation.
In general, many possible terms are generated and appear in $S_{\rm eff}$.
Each possible term defines a coupling of the theory.
Each coupling can be classified as marginal (the size of the coupling is unchanged by the RG flow), 
relevant (the coupling grows under the flow) or irrelevant (the coupling goes to zero under the flow).
It is a dramatic insight of Wilson that almost all couplings in any given quantum field theory are irrelevant and 
so the low energy theory is characterized by a handful of parameters.

To compare how parameters have changed, we want to compare the action before and after integrating out a shell of modes.
For this comparison to be meaningful, the integration domains, before and after integrating the shell out, must match.
To achieve that we rescale momenta ($k$), positions ($x$) and the field ($\phi$) as follows
\bea
\vec{k}\to\vec{k}'&=&{\vec{k}\over 1-\epsilon}\cr
\vec{x}\to\vec{x}'&=&(1-\epsilon)\vec{x}\cr
\vec{\phi}\to\vec{\phi}'&=&{\vec{\phi}\over 1-\epsilon}\label{rescale}
\eea
The rule for scaling momenta is fixed by matching integration domains.
Since momentum is an inverse length, positions scale oppositely to momenta.
The rule for how the field is scaled ensure that the free field theory, described by the action
\bea
S=\int d^4 x {1\over 2}\vec{\nabla}\phi\cdot\vec{\nabla}\phi
\eea
is invariant under the renormalization group flow.
Here we have committed to $d=4$ dimensions.
The scaling of the field depends on $d$.

So there are two reasons why the couplings change: (i) we integrate over high momentum modes in (\ref{outershell})and 
(ii) we rescale according to (\ref{rescale}).
If we are very close to the free field fixed point couplings are weak and we can neglect the effect coming from doing the
integral in (\ref{outershell}).
In this case we simply discard the high momentum modes and use the scaling (\ref{rescale}).
With the above transformation, a general term will scale as follows
\bea
\lambda \phi^n (\vec{\nabla}\phi\cdot\vec{\nabla}\phi)^m \to
(1-\epsilon)^{n+4m-4}\,\lambda\, \phi^{\prime n} (\vec{\nabla}'\phi'\cdot\vec{\nabla}'\phi')^m\cr
\eea
so that the coupling changes as
\bea
\lambda\to \lambda'=(1-\epsilon)^{n+4m-4}\,\lambda
\eea
Since $1-\epsilon <1$, the only relevant terms in the Lagrangian are
\bea
\phi,\qquad\phi^2,\qquad\phi^3
\eea
The only marginal terms are\footnote{Quantum corrections, that is the corrections from actually integrating the shell of
high momentum modes, make this term irrelevant. We call it marginally irrelevant.}
\bea
\phi^4,\qquad \vec{\nabla}\phi\cdot\vec{\nabla}\phi
\eea
There are an infinite number of irrelevant terms
\bea
\phi^{n+4},\qquad \phi^n (\vec{\nabla}\phi\cdot\vec{\nabla}\phi)^m
\qquad m,n\ge 1
\eea
Thus, although there are an infinite number of possible couplings in the microscopic short distance description, only five
couplings in total survive in the long distance theory!

\noindent
{\bf Key Idea:} Near the free field fixed point, the renormalization group transformation simply discards the high momentum
modes of the field that is being coarse grained.
This is precisely what the trained deep network does.

\subsection{Block Spin Transformations}\label{sec:blockspin}

The implementation of the renormalization group for the RBM trained on Ising data is block spin transformation as
discussed in Section \ref{ising}.
Large groups of spins are averaged to produce the block spin.
Any fluctuations of the spins on the lattice, that average to zero and are contained within the block being averaged, 
will be removed by the coarse graining.
This implies that irrelevant perturbations are the ones that produce localized changes to the state, with the changes
averaging to zero.
This is in line with our intuition: irrelevant perturbations are associated with short length scales.

\section{RGM}\label{RGM}

The connection between deep learning and the renormalization group that we are arguing for in this study suggests
economic parameterizations of the weight matrix.
Understanding these parametrizations will shed light on the generalization puzzle.
Since the specific form adopted for the weights is motivated by the renormalization group, we refer to the resulting deep
network as a renormalization group machine, or RGM for short.

The economy of the parametrization follows because we know that only the big singular values count and that these 
only have low Fourier modes.
We are going to build this into the weight matrix.

Using the singular value decomposition, we can write the weight matrix $W_{(m,a)}$ as
\begin{equation}
W_{ma} = \sum_{I=1}^{N_{h}}S_{I} | v, I \rangle_{m} \langle h, I |_{a}
\end{equation}
where $m = 1, ..., N_{v}=L_v^2$ and $a = 1, ..., N_{h}=L_h^2$.
We have in mind input data given by two dimensional Ising lattices that have $N_v$ sites arranged in an $L_v\times L_v$
square, or images that have a total of $N_v$ pixels.
The data output by the network are states of a two dimensional Ising lattice with $N_h$ sites, or images with a total of
$N_h$ pixels.
We will use the language ``visible lattice'' and ``hidden lattice'' for both Ising and images in what follows.
The index $m$ runs over the sites of the visible lattice, so we can trade it for the coordinates $(l,n)$ of a site in the 
visible lattice. 
Here $l$ and $n$ both run from 1 to $L_v$.
Similarly, the index $a$ runs over the sites of the hidden lattice, so we can trade it for the coordinates $(b,c)$ of a site
in the hidden lattice. Here $b$ and $c$ both run from 1 to $L_h$.
With the new notation we can write the weight matrix $W_{ma}$ as
$$
W_{ma}\to W_{(l,n)\,(b,c)} = \sum_{I=1}^{N_h}S_{I}|v,I\rangle_{(l,n)}\langle h, I |_{(b,c)}
$$
Using the known simplifications we have
$$
W_{(l,n)\, (b,c)} = \sum_{I=1}^{\kappa}S_{I}|v,I\rangle_{(l,n)}\langle h, I |_{(b,c)}
$$
where $\kappa$ is the number of large singular values and typically $\kappa \ll N_h$.
We can also simplify the singular vectors appearing in the above expansion.
To do this we use the discrete Fourier transform to write
\begin{equation}
| v,I\rangle_{(k,p)}=\sum_{m=1}^{L_{v}}\sum_{n=1}^{L_{v}}|v,I\rangle_{(m,n)}e^{-i2\pi (\frac{lk}{L_{v}}+\frac{np}{L_{v}})}
\end{equation}
for the visible vector and
\begin{equation}
| h,I\rangle_{(x,y)}=\sum_{a=1}^{L_{h}}\sum_{b=1}^{L_{h}}\langle h,I|_{(a,b)}e^{-i2\pi (\frac{ax}{L_{h}}+\frac{by}{L_{h}})}
\end{equation}
for the hidden vector.
The inverse Fourier transform is
\begin{equation}
|v, I \rangle_{(l,n)} = \frac{1}{{L_{v}}^{2}}\sum_{k=1}^{L_{v}}\sum_{p=1}^{L_{v}}|v,I\rangle_{(k,p)}
e^{i2\pi(\frac{lk}{L_{v}}+\frac{np}{L_{v}})}
\end{equation}
\begin{equation}
| h,I\rangle_{(a,b)}=\frac{1}{{L_{h}}^{2}} \sum_{x=1}^{L_{h}}\sum_{y=1}^{L_{h}}\langle h,I|_{(x,y)}e^{i2\pi (\frac{ax}{L_{h}}+\frac{by}{L_{h}})}
\end{equation}
The coarse graining interpretation implies that the singular vectors only have support at low Fourier modes.
Consequently a more efficient parametrization of the visible singular vector is
\begin{equation}
|v,I\rangle_{(l,n)} = \frac{1}{{L_{v}}^2} \sum_{k=1}^{\alpha}\sum_{p=1}^{\alpha} C^{I}_{(k,p)}e^{i2\pi(\frac{lk}{L_{v}}+\frac{np}{L_{v}})}
\end{equation}
to form the visible vector $|v\rangle$.
Typically  $\alpha \ll L_h<L_v$.
For the hidden vector, $\langle h|$, we use
\begin{equation}
\langle h,I |_{(a,b)}=\frac{1}{{L_{h}}^2}\sum_{x=1}^{\alpha}\sum_{y=1}^{\alpha}\frac{C^{I}_{(x,y)}}{2}
e^{-i2\pi(\frac{ax}{L_{h}}+\frac{by}{L_{h}})}
\end{equation}
The fact that the same Fourier coefficients appear in the expansion of both the visible and the hidden singular vectors
is a consequence of the fact that the network is performing a renormalization group transformation, so that the low
frequency modes of the two vectors agree.
Further, the fact that the singular vectors are real implies that
\bea
C^{I}_{(k,p)}=C^{I}_{(-k,-p)}
\eea
Since the singular vector $S_I$ appears as an overall factor, it can be absorbed into the definition of $C^{I}_{(k,p)}$.
We will not do this, but it is worth noting since this observation implies that the $S^I$ should not be counted when the 
number of effective parameters is counted.
Thus, we finally obtain
\begin{eqnarray}
W_{(m,n)\, (a,b)} = \frac{1}{{L_{v}}^2} \frac{1}{{L_{h}}^2} \sum_{I=1}^{\kappa}  \sum_{k,p=1}^{\alpha}
\sum_{x,y=1}^{\alpha}\cr
S_I \frac{C^{I}_{(k,p)}C^{I}_{(x,y)}}{2}e^{i2\pi\left( (\frac{mk}{L_{v}}+\frac{np}{L_{v}})-(\frac{ax}{L_{h}}+\frac{by}{L_{h}})\right)}
\end{eqnarray}
The only unknowns in the above formula for the weight matrix are the coefficients $C^{I}_{(k,p)}$ and the singular 
values $S^I$. 
To obtain these coefficients we compute the data covariance matrix for the training data set.
The eigenvectors associated with the largest eigenvalues are identified with the visible singular vectors
$|v,I\rangle_{(l,n)}$. 
By taking a Fourier transform of this vector and setting small Fourier modes to zero, we obtain the $C^{I}_{(k,p)}$.

Although we do not have a perfect formula for the singular values $S_I$, we have developed a simple rule that works
reasonably well in practice.
While the data determines the singular vectors, the singular values are determined by the coarse graining rule. 
Large singular values correspond to relevant modes and small singular values to irrelevant modes.
To estimate the size of the singular value we estimate the magnitude of the vector 
\bea
|I\rangle_{(r,s)}=\sum_{l,n=1}^{L_v}C_{(r,s)\, (l,n)}|v,I\rangle_{(l,n)}
\eea
where $C_{(r,s)\, (l,n)}$ is a block spin transformation and $|v,I\rangle_{(l,n)}$ is a normalized singular vector.

The hidden and visible biases are chosen to be equal to the hidden and visible singular vectors associated to the largest singular
vector.
This choice is intuitively appealing: they bias configurations towards the subspace on which the data is concentrated.
In addition, the trained biases are in remarkably good agreement with the singular vectors associated with the largest singular
vectors.

The weakest part of our discussion has been in fixing the overall scales of the biases and the term involving the weights.
Thanks to the activation functions in the network, much of the sensitivity to these scales is removed.

\section{Ising model}\label{sec:ising}

The Ising model Hamiltonian is
\begin{equation}
  H_{I} = -J\sum_{\corr{{\bf i},{\bf j}}}s_{\bf i}s_{\bf j}
  \label{eq:ising}
\end{equation}
where the sum runs over nearest neighbors.
In this study we take the spins to sit at the sites of a rectangular lattice.
The probability distribution implied by the Hamiltonian is $e^{-\beta H}$ so that $\beta$ and $J$ are not
independent parameters. 
We set $J=1$ and use the temperature $T$ to set $\beta={1\over T}$.
By employing Monte Carlo methods we generate states of this 2D Ising lattice model.
The training data set used to train the networks described in Section \ref{ising} is comprised of 40 000 samples of this
lattice model.
Each training epoch used the complete data set, and training was completed is roughly 6000 epochs.

\section{Overlapping Coarse Graining}\label{blockoverlap}

We can define many different coarse graining rules that use block spin transformations.
Many of these rules result in overlap between different averaging blocks.
The coarse graining rules at work in the trained RBM network is a quantitative match for block spin transformations 
with overlapping blocks, so it is important to understand the effect that overlap has if we are to understand what 
is happening in the network.

\subsection{Generating the weight matrix}

We create an effective weight matrix that, when applied to an input vector, will apply the coarse graining rule.
The input vector is a 6400 dimensional vector, obtained by rearranging the elements of an $80\times 80$ matrix.
Each column of the weight matrix is constructed to average a single block of spins on the lattice.
Thus, each column only has entries of $1$'s and $0$'s.
For all coarse graining rules we move by two elements in the x and y direction for each new averaging block.
This implies that the averaged vector is a 1600 dimensional vector.

The first block averaged is centered at the topmost and leftmost corner of the lattice.
Any elements that should be averaged in the block, but do not exist in the input matrix, are discarded.
This occurs for blocks averaging the edges of the input matrix.

\subsection{Effect of overlapping }

Weight matrices generated with block spins that have no overlap, have no variation in their singular values.
For example, the weight matrix generated using an averaging block of size 2x2 has all singular values equal to 2.
Each element in the output vector is obtained by averaging 4 elements from the input vector and each element in the
input vector contributes to a unique element in the output vector.
The hidden and visible singular vectors of the weight matrix are identical.

\begin{figure}[h!]
\centering
\includegraphics[scale=0.5]{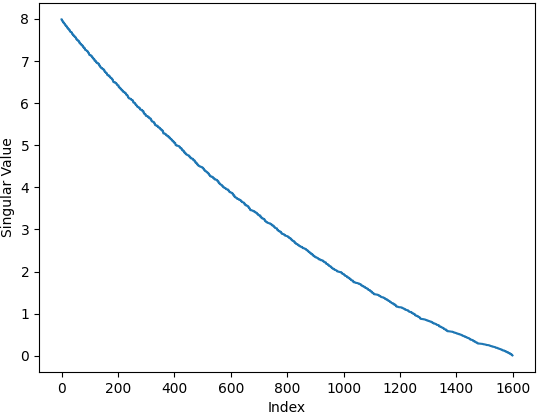}
\caption{Singular values from a weight matrix generated from a coarse graining rule with little overlap.}\label{small_overlap_s}
\end{figure}

As overlap begins to occur, we see some variation in the singular values.
The weight matrix used to plot figure \ref{small_overlap_s} has a coarse grain rule that uses averaging blocks with size 
$4\times 4$, so that 16 spins are averaged to produce a single element in the output vector. 
In this case, the singular values decay almost linearly, as in Figure \ref{small_overlap_s}.
Since the block is moved by two elements for each new average, a single spin contributes to multiple outputs.

\begin{figure}[h!]
\centering
\includegraphics[scale=0.4]{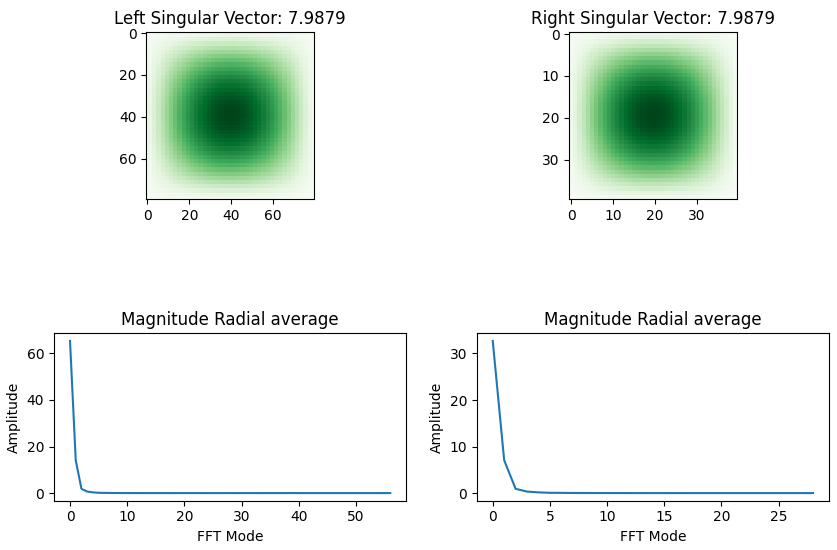}
\caption{Singular vectors from a weight matrix generated from a coarse graining rule with little overlap for large singular values.}\label{small_overlap_vectors_big_s}
\end{figure}

\begin{figure}[h!]
\centering
\includegraphics[scale=0.4]{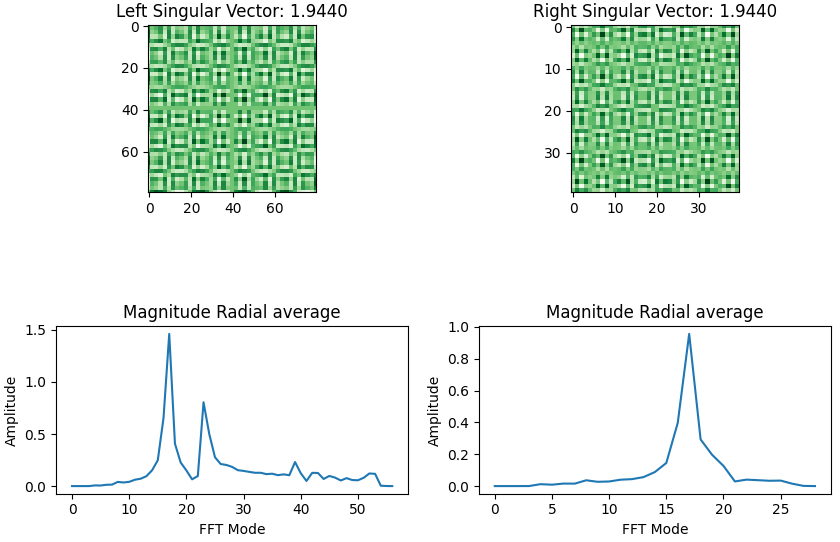}
\caption{Singular vectors from a weight matrix generated from a coarse graining rule with little overlap for small singular values.}\label{small_overlap_vectors_small_s}
\end{figure}

Singular vectors corresponding to large singular values, shown in Figure \ref{small_overlap_vectors_big_s}, 
are concentrated at low frequencies.
Additionally, the visible and hidden singular vectors have the same shape.
The hidden singular vectors agree perfectly with the visible singular vectors after rescaling by a factor of 2.
For small singular values, shown in Figure \ref{small_overlap_vectors_small_s}, the hidden and visible singular vectors 
no longer agree and they are no longer concentrated at low frequencies.

\begin{figure}[h!]
\centering
\includegraphics[scale=0.5]{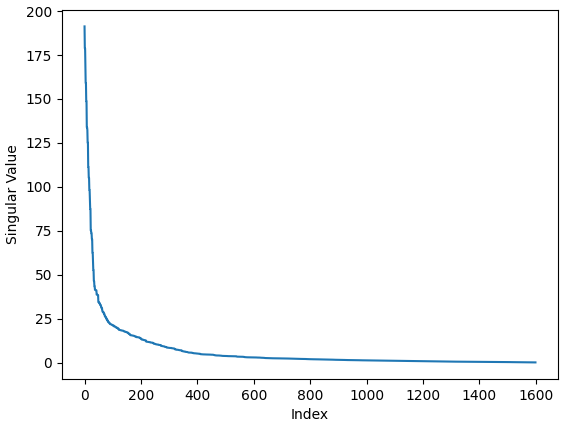}
\caption{Singular values from a weight matrix generated from a coarse graining rule with large overlap.}\label{large_overlap_s}
\end{figure}

As the overlap between the averaging blocks grows, we see a small number of very large singular values that fall off rapidly.
Figure \ref{large_overlap_s} is from a weight matrix with a block size of 20x20, averaging 400 elements for each output.
The large block size means each element will contribute to a large number of output elements and so has a significant 
overlap between averaging blocks.

For large singular values, the Fourier transform of the visible and hidden singular vectors are again identical and they
are again concentrated at low frequencies.
The singular vectors associated to small singular values, are again not concentrated at low frequencies and the Fourier
transform of the visible and hidden singular vectors no longer agree.

\section{Radial FFT}\label{radialFFT}

We study the frequency domain of our singular vectors using the two-dimensional discrete Fourier transform\footnote{Note
that the conventions for the Fourier transform in this section are related to those of Section \ref{RGM} by shifting the 
momentum space lattice.} 
\begin{equation}\label{fft2d}
F(u,v)=\sum_{m=-{L_v\over 2}}^{L_v\over 2}\sum_{n=-{L_h\over 2}}^{L_h\over 2}f(v_{m},h_{n})
e^{-i2\pi (\frac{mu}{L_{v}}+\frac{nv}{L_{h}})}
\end{equation}
Assuming that we want to learn features of the data that are rotationally invariant, it makes sense to consider the radially
averaged Fourier transform.
This significantly simplifies the presentation of the data and it makes comparisons easier to interpret. 

After we have completed the singular value decomposition of the weight matrix, we have the hidden and visible singular 
vectors as one-dimensional vectors.
These one-dimensional vectors are rearranged to give states of a two-dimension square lattice with $L_{v}\times L_{v}$ lattice
sites for the visible vectors and $L_{h}\times L_{h}$ sites for the hidden vectors.
The center of the lattice is the zero frequency and the frequency increases radially from the center.
We average values in an annulus centered on the zero frequency site, to get a single radially averaged frequency mode.

\section{Numerical results}\label{sec:numeric}

In this Appendix we collect the numerical results that were briefly mentioned in the body of the article.

\subsection{Deep RBM trained on Ising Data}

In Section \ref{ising}, we claimed that each layer of the layered RBM performs a renormalization group transformation.
The numerical evidence for this statement again comes from looking at the Fourier transform of singular vectors associated
to large singular values. 
We again find that these singular vectors have support on low Fourier modes and that the hidden singular vector is obtained
from the visible singular vector by discarding high Fourier modes.

\begin{figure}[h]
    \centering
    \includegraphics[width=0.4\textwidth]{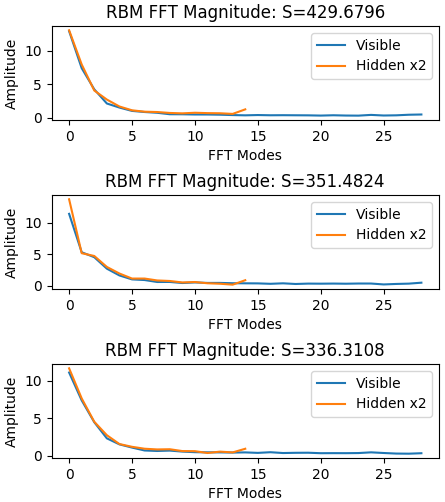}
    \caption{The radial Fourier transform of singular vectors of the trained weight matrix, corresponding to large singular
values, for an RBM trained on Ising data. The singular vectors shown were obtained from a singular value decomposition
of the weight matrix associated to the second layer of a three layer network.}
    \label{fig:ising2layerlarge}
\end{figure}

\begin{figure}[h]
    \centering
    \includegraphics[width=0.4\textwidth]{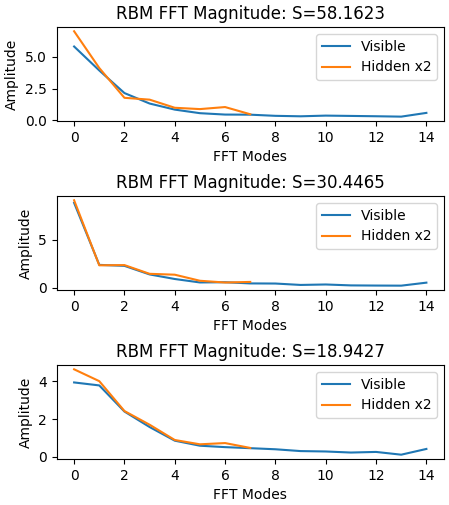}
    \caption{The radial Fourier transform of singular vectors of the trained weight matrix, corresponding to large singular
values, for an RBM trained on Ising data. The singular vectors shown were obtained from a singular value decomposition
of the weight matrix associated to the third layer of a three layer network.}
    \label{fig:ising3layerlarge}
\end{figure}

Figure \ref{fig:ising2layerlarge} shows the radially averaged Fourier transform for the large singular vectors of layer 2,
while Figure \ref{fig:ising3layerlarge} shows the same thing for layer 3.


\subsection{MNIST}

The MNIST data set is a rather simple data set.
We have seen that training from the RGM cuts training time by a factor of 4.
The fashion MNIST data set, was introduced as a more challenging data set.
We have compared the performance of the RGM, a trained RBM and a trained RGM on the fashion MNIST data set.
In this case, we find a speed up of 10 times.
Recall that the flower data set used in Section \ref{flowers} was even more challenging.
It lead to a speed up of about 100 times.
The fashion MNIST results are shown in Table \ref{table:fashion}.

\vfill
\pagebreak

\begin{table}[ht]
\caption{Results of the RBM and RGM on the fashion-MNIST dataset. The RBM is trained for 1000 epochs and the RGM has been trained for 100 epochs.}\label{table:fashion}
\centering
\begin{tabular}{|c|c|c|c|}
\hline
Original Image & RBM & RGM & Trained RGM \\ \hline &&&\\
\includegraphics[scale=0.1]{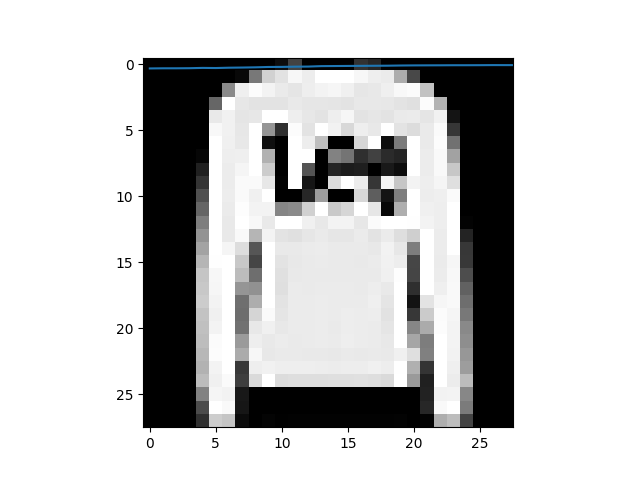}&\includegraphics[scale=0.1]{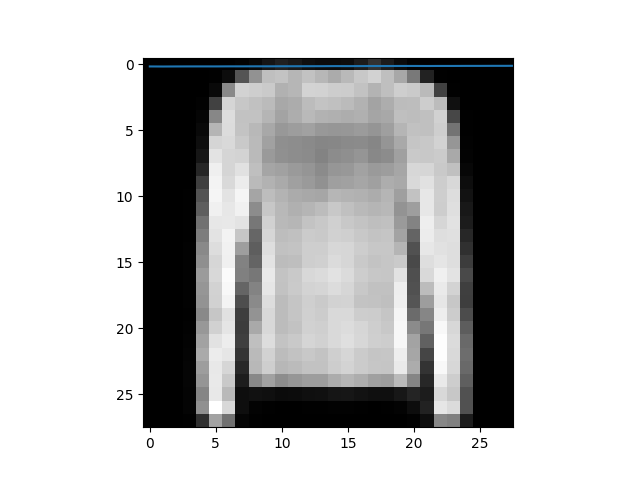}&\includegraphics[scale=0.1]{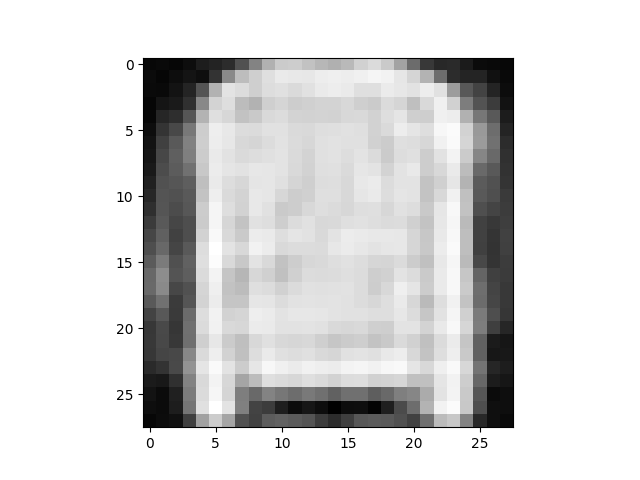}&\includegraphics[scale=0.1]{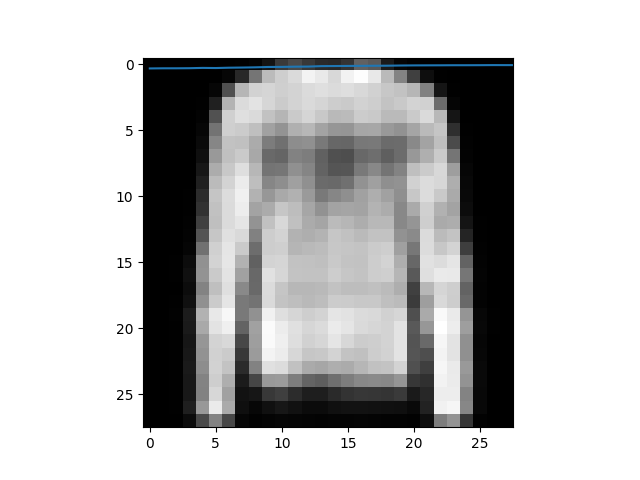}\\ 
\includegraphics[scale=0.1]{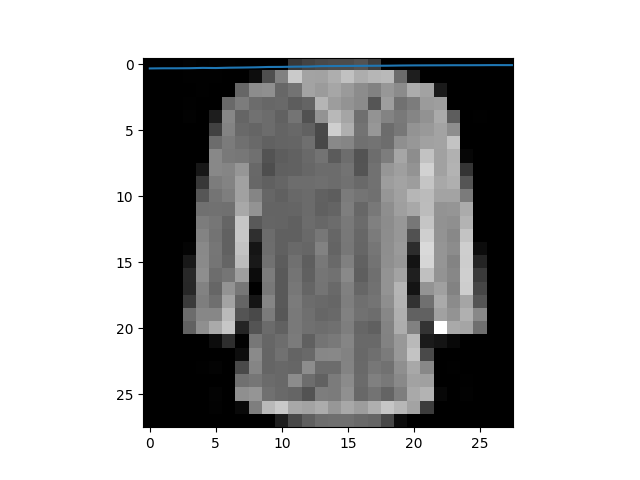}&\includegraphics[scale=0.1]{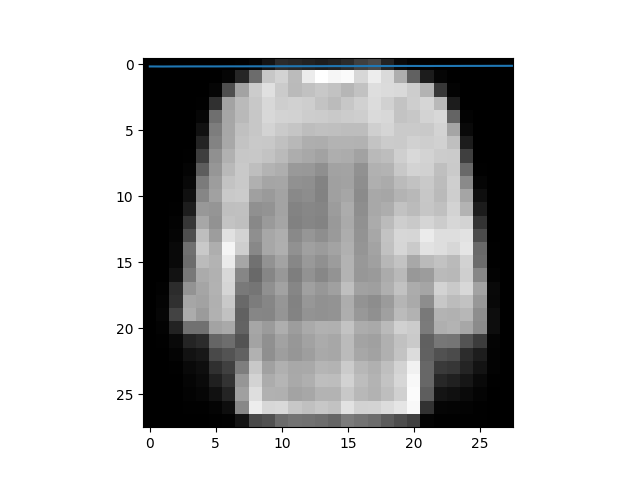}&\includegraphics[scale=0.1]{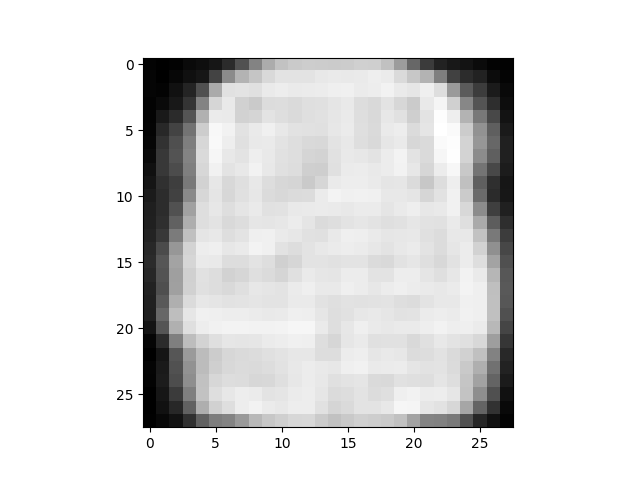}&\includegraphics[scale=0.1]{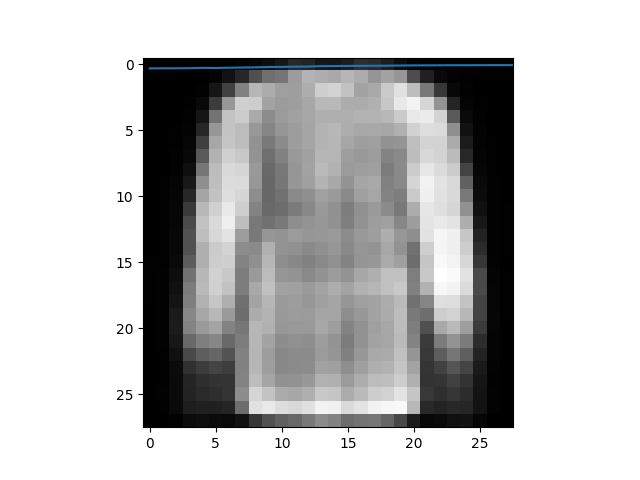} \\
\includegraphics[scale=0.1]{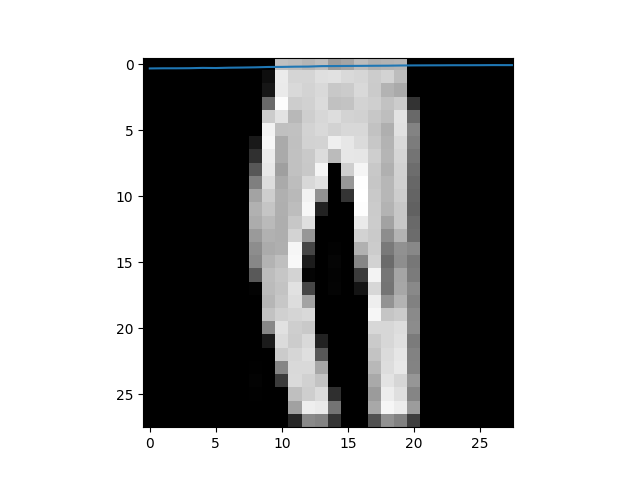}&\includegraphics[scale=0.1]{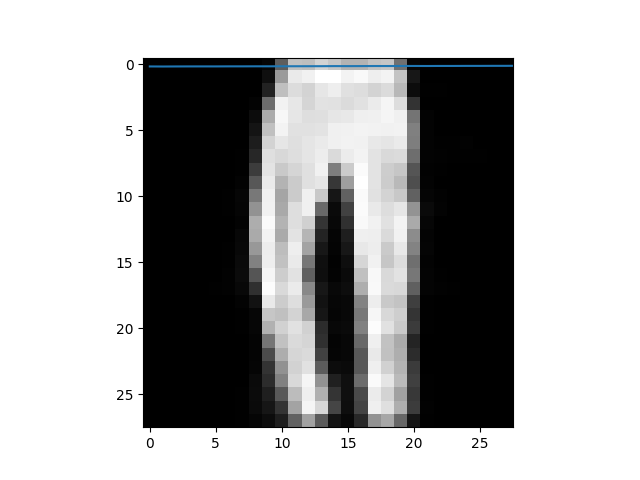}&\includegraphics[scale=0.1]{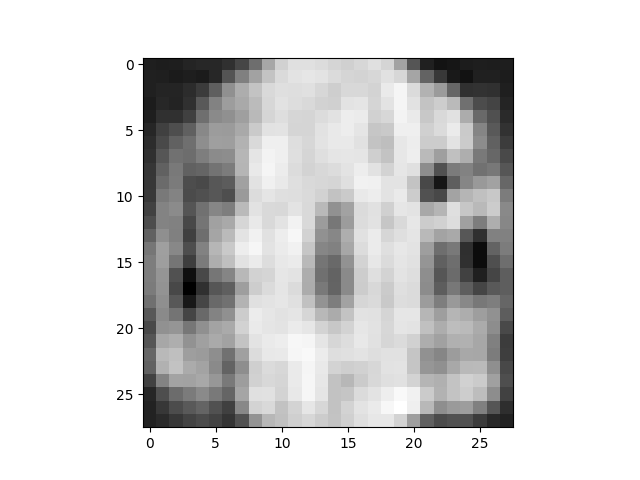}&\includegraphics[scale=0.1]{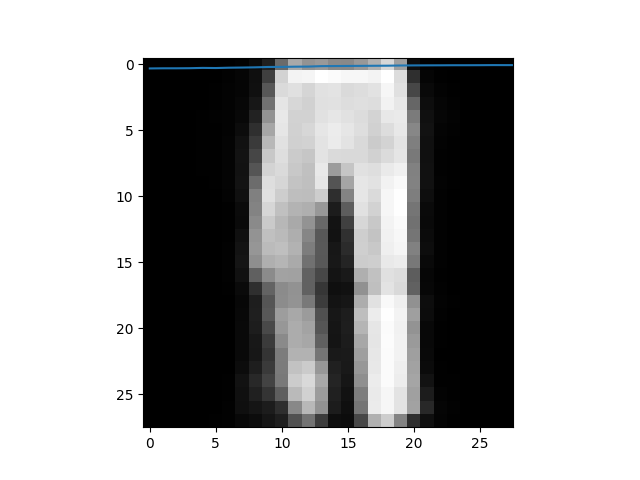}\\ 
\includegraphics[scale=0.1]{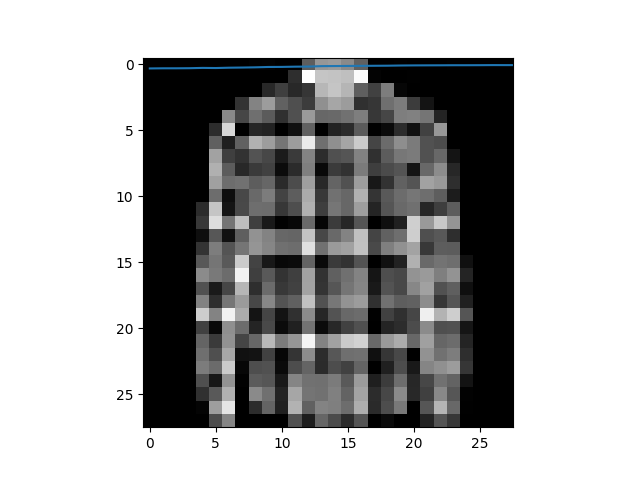}&\includegraphics[scale=0.1]{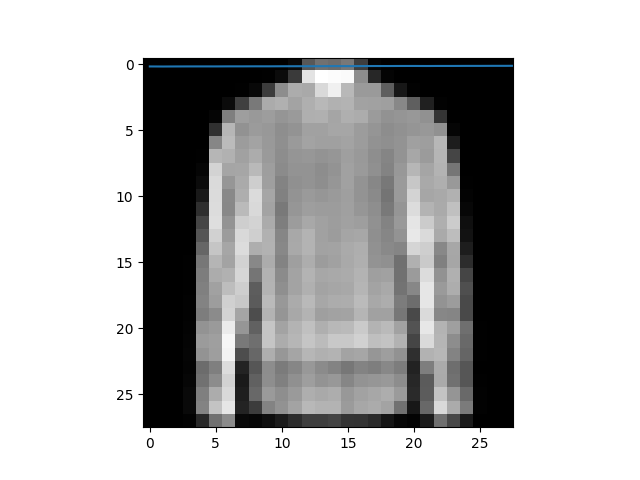}&\includegraphics[scale=0.1]{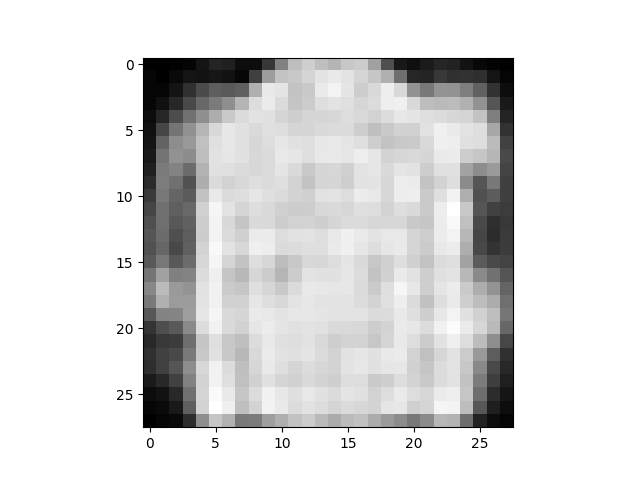}&\includegraphics[scale=0.1]{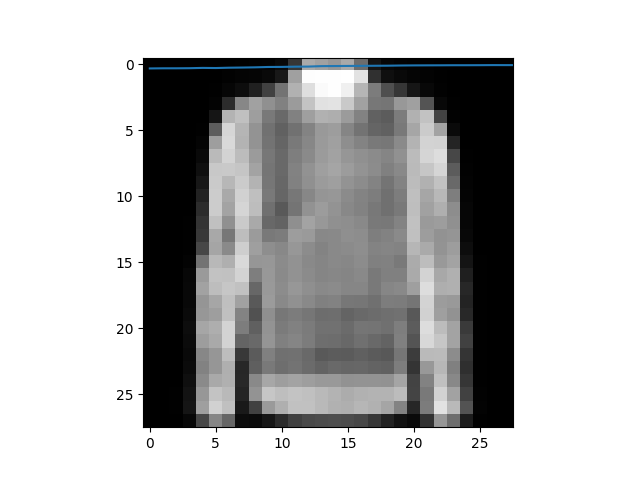} \\
\hline
\end{tabular}
\end{table}

\subsection{Singular value decomposition of the block spin transformation coarse graining}

\begin{figure}[h!]
    \centering
    \includegraphics[width=0.4\textwidth]{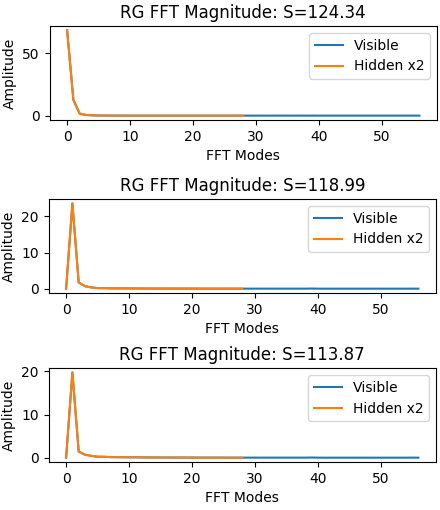}
    \caption{The Fourier transform of the visible and hidden singular vectors associated to the five largest singular vectors,
                obtained from the singular value decomposition of the block spin averaging matrix.}
    \label{fig:rgsingularvectors}
\end{figure}

\begin{figure}[h!]
    \centering
    \includegraphics[width=0.4\textwidth]{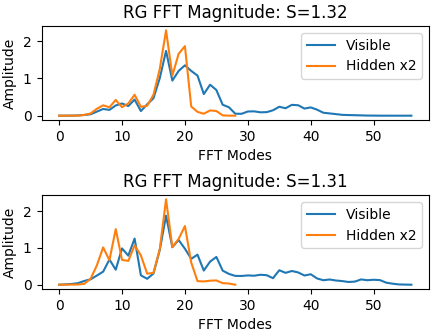}
    \caption{The Fourier transform of the visible and hidden singular vectors associated to the five largest singular vectors,
                obtained from the singular value decomposition of the block spin averaging matrix.}
    \label{fig:smallrgsingularvectors}
\end{figure}

We have used both the block spin transformation to coarse grain and the momentum space renormalization group.
In this Appendix we show the equivalence of the two by performing a singular value decomposition of the matrix 
implementing the block spin coarse graining.
The results are given in Figure \ref{fig:rgsingularvectors} and Figure \ref{fig:smallrgsingularvectors}.
The visible and hidden singular vectors have all of their support at low Fourier modes, and the hidden singular vector
is again obtained from the visible singular vector by discarding large Fourier modes.
In this case too, singular vectors associated to small singular values are again spread across Fourier space and 
the hidden and visible singular vectors again no longer agree.

\bibliography{refs}

\end{document}